\pdfoutput=1

\documentclass[a4paper,fleqn]{cas-sc}
\usepackage{subfigure}
\usepackage{times}
\usepackage{amsmath}
\usepackage{amsfonts}
\usepackage{float}
\usepackage{gensymb}
\usepackage{algorithm}
\usepackage{algpseudocode}
\usepackage{array}
\usepackage{textcomp}
\usepackage{stfloats}
\usepackage{url}
\usepackage{verbatim}
\usepackage{graphicx}  
\usepackage{pifont}
\usepackage{float}
\usepackage{enumitem}
\usepackage{caption}
\usepackage{multirow} 
\usepackage{longtable} 
\usepackage{array} 
\usepackage{makecell}
\usepackage{threeparttable} 
\usepackage{booktabs}
\usepackage{listings}
\usepackage{xcolor}

\usepackage[numbers]{natbib}
\usepackage{amssymb}

\usepackage[numbers]{natbib}
\usepackage{amssymb}
\def\tsc#1{\csdef{#1}{\textsc{\lowercase{#1}}\xspace}}
\tsc{WGM}
\tsc{QE}

\begin{document}

\let\WriteBookmarks\relax
\def\floatpagepagefraction{1}
\def\textpagefraction{.001}

\shorttitle{Distillation Enhanced Time Series Forecasting Network with Momentum Contrastive Learning}
\shortauthors{Qianqian Ren et~al.}
\title [mode = title]{Distillation Enhanced Time Series Forecasting Network with Momentum Contrastive Learning}  

\author[1]{Haozhi Gao}
\ead{2221942@s.hlju.edu.cn}

\author[1]{Qianqian Ren}  
\cormark[1]
\ead{renqianqian@hlju.edu.cn}
\address[1]{Department of Computer Science and Technology, Heilongjiang University, Harbin, 150080, China}

\author[2]{Jinbao Li}
\cormark[1]
\ead{lijinb@sdas.org}
\address[2]{Shandong Artificial Intelligence Institute,School
of Mathematics and Statistics, Qilu University of Technology, Jinan 250014, China}

\cortext[1]{Corresponding author.}

\begin{abstract}
Contrastive representation learning is crucial in time series analysis as it alleviates the issue of data noise and incompleteness as well as  sparsity of supervision signal. However, existing constrastive learning frameworks usually focus on intral-temporal features, which fails to fully exploit the intricate nature of time series data. To address this issue, we propose DE-TSMCL, an innovative distillation enhanced framework for long sequence time series forecasting.
Specifically, we design a learnable data augmentation mechanism which adaptively learns whether to mask a timestamp to obtain optimized sub-sequences. Then, we propose a contrastive learning task with momentum update to explore inter-sample and intra-temporal correlations of time series to learn the underlying structure feature on the unlabeled time series.
Meanwhile, we design a supervised task  to learn more robust representations and facilitate the contrastive learning process. Finally, we jointly optimize the above two tasks. By developing model loss from multiple tasks, we can learn effective representations for downstream forecasting task.
Extensive experiments, in comparison with state-of-the-arts, well demonstrate the effectiveness of DE-TSMCL, where the maximum improvement can reach to 27.3\%. Source code for the algorithm is available at https://github.com/gaohaozhi/DE-TSMCL.

\end{abstract}

\maketitle
\section{Introduction}
Time series forecasting plays a critical role in various domains, including  finance, economics, weather forecasting, and resource management. Accurately predicting future values based on historical data is essential for decision-making and planning purposes in these fields\cite{li2023mts,ozyurt2022contrastive,rasul2023lag}. For the current time series forecasting study, the critical problem is how to leverage the inherent structure information of time series to learn discriminate representations to achieve better forecasting accuracy\cite{tonekaboni2021unsupervised,yue2022ts2vec}. The ability of a model to learn these representations is critical for enhancing its performance\cite{xiao2024densely, franceschi2019unsupervised}. 
The primary focus of our work in this article is to develop an efficient framework that integrates representation learning and forecasting in a unified manner.


With the emergence of contrastive learning methods, self-supervised representation learning and time series prediction have witnessed increased attention and progress. Several notable methods such as TS2Vec\cite{yue2022ts2vec}, 
TS-TCC \cite{eldele2021time}, TF-C \cite{zhang2022self}, and CPD \cite{deldari2021time} have been developed in this domain. The objective of contrastive learning is to train one or multiple encoders that can generate representations where similar instances are closer to each other, while dissimilar instances are farther apart.
In particular, Yue et al. \cite{yue2022ts2vec} introduce a comprehensive framework called TS2Vec for learning time series representations. This framework utilizes a hierarchical approach to distinguish between positive and negative samples.
Eldele et al. \cite{eldele2021time} propose a time-series representation learning framework called TS-TCC, which employs temporal and contextual contrastive learning to extract discriminative representations.
Another notable contrastive learning approach, CoST, is introduced by Woo et al. \cite{woo2022cost} specifically for predicting long sequence time series. It transforms the data into the frequency domain to reveal the seasonal representations of the sequence, enabling a comprehensive understanding of its characteristics.
These contrastive learning methods have demonstrated their effectiveness in learning informative representations for time series data. 

Despite the promising results in time series forecasting, there are three crucial aspects that have often been neglected. These aspects have a significant impact on the accuracy and generalization ability of the models in real-world scenarios:

\begin{itemize} 
\item \textbf{Data Noise and Distribution Shift:} Real-world time series data often suffers from noise and incompleteness due to various environmental factors. This introduces additional challenges in accurate forecasting as the noise can corrupt the underlying patterns and relationships within the data. Moreover, the distribution of real-world data may differ from the training data, leading to a distribution shift. This inconsistency can result in poor model performance and inaccurate predictions. 

\item \textbf{Dependence on Single Salient Feature:}
Dependence on a single salient feature in time series forecasting can indeed limit the model's ability to capture the full complexity of the data and generalize well to new instances. Over-reliance on a single feature may lead to the model learning shortcuts or missing out on other relevant predictive features, reducing its performance and interpretability. Additionally, existing models often treat multivariate time series as a single integrated representation, disregarding the correlations between individual instances or variables within the series. 

\item \textbf{False Positive Focusing in Contrastive Learning:} 
Contrastive learning approaches typically rely on generating positive pairs and negative patterns based on prior knowledge or strong assumptions about the data distribution, which makes the model pay excessive attention on the distances between positive and negative sample pairs within the sample space. This can lead to a neglect of the similarity between different overlapping subsequences within the same sequence, thereby limiting the model's ability to capture the full complexity of the data and potentially impacting its performance.
\end{itemize}

To address the above challenges, we develop {DE-TSMCL}, a  distillation enhanced framework for time series forecasting based on momentum contrastive learning, which contains three key components: \textbf{learnable data augmentation}, \textbf{distillation enhanced representation} and \textbf{momentum constrastive learning}.
First, we propose learnable data augmentation to learn whether to mask a sample to transform the original two overlapping subseries into enhanced views, which will be fed into the following teacher and student network and jointly optimized with the downstream forecasting task in an end-to-end fashion. 
Second, we introduce knowledge distillation technique into representation learning process. We simultaneously train two models, the teacher and the student. The teacher makes full use of of all available knowledge to capture the temporal features of time series to achieve better representation performance. 
Third, in order to calculate the similarity of samples from different timestamp,
supervised task is considered. At the same time, considering the commonalities between multiple samples and effectively alleviating the
problems of data sparsity, we design a self-supervised strategy powered with momentum contrastive training to further boost the performance. Finally, we jointly optimize these two tasks.

The contributions of this thesis can be summarized as follows.

\begin{itemize}
\item We propose a Distillation Enhanced Time Series Forecasting Network via Momentum Contrastive Learning (DE-TSMCL) to improve the time series forecasting performance. To the best of our knowledge, we are the first to apply KD techniques among models that rely on different overlapping subseries in the time series forecasting task.

\item We design supervised and self-supervised learning task for time series forecasting, moreover, we jointly optimize them. In addition, we apply momentum contrastive training to update the teacher and student network at different scales, which solves the noisy and inconsistent issue due to the back-propagation. 
\item Extensive experiments on four benchmark datasets from different domains demonstrate that our model advances the forecasting performance compared
with other baselines 

\end{itemize}

The paper is structured as follows: Section \ref{2} summarizes the related work. Section \ref{3} gives the preliminaries. Section \ref{4} introduces our proposed {DE-TSMCL model}. Section \ref{5} gives the comparison and evaluation results of the proposed model on four real-world datasets. Finally, the paper is concluded in Section \ref{6}.

\section{Related Work}\label{2}
In this section, we briefly review some related work on time series forecasting, contrastive learning and knowledge distillation.

\subsection{Time Series Forecasting}
Time series forecasting has been extensively studied, and various models have been developed to tackle this task. Traditional approaches include autoregressive models such as ARIMA (AutoRegressive Integrated Moving Average)\cite{zhang2003time}, exponential smoothing methods like Holt-Winters, and state space models like Kalman filters. These models capture temporal dependencies and exploit statistical properties of time series data.

In recent years, deep learning models have been widely used in time series forecasting due to their ability to automatically learn complex patterns and dependencies. Convolution Neural networks(CNNs)\cite{wan2019multivariate,bai2018empirical}, Recurrent Neural Networks (RNNs)\cite{wen2017multi,oreshkin2019n,salinas2020deepar}, particularly variants like Long Short-Term Memory (LSTM) and Gated Recurrent Units (GRU), have shown success in capturing long-term dependencies in time series data. Temporal Convolution Network (TCN\cite{bai2018empirical}) introduces dilated convolutions for time series forecasting and demonstrates their superior efficiency and prediction performance compared to RNNs\cite{bai2018empirical}. Moreover, LSTNet \cite{lai2018modeling} integrates CNNs and RNNs to simultaneously capture short- and long-term dependencies. 

In addition, attention mechanisms have been integrated into RNN-based models to improve the model's forecasting accuracy\cite{li2019enhancing,zhou2021informer,wu2021autoformer,zhou2022fedformer}. Transformer-based architectures, originally designed for natural language processing tasks, have also been adapted for time series forecasting by employing self-attention mechanisms.
LogTrans\cite{li2019enhancing} employs convolutional self-attention layers with a LogSparse design to capture local information features. Informer\cite{zhou2021informer} introduces a ProbSparse self-attention mechanism that uses distillation techniques to efficiently extract the most crucial keys. Autoformer \cite{wu2021autoformer} incorporates the concepts of decomposition and auto-correlation from traditional time series analysis methods. FEDformer \cite{zhou2022fedformer} employs a Fourier-transform structure and achieve linear complexity. 


\subsection{Contrastive Learning}
In recent years, there has been growing interest in applying contrastive learning for time series forecasting. 
Oord et. al propose contrastive predictive coding (CPC)\cite{oord2018representation} to predict the subsequent latent variable, as opposed to negative samples drawn from the proposed distribution.
Temporal and contextual contrasting (TS-TCC)\cite{eldele2021time}, a variant of CPC\cite{oord2018representation}, aims to optimize the concordance between robust and subtle enhancements of the same instance within an autoregressive framework.
TS2Vec\cite{yue2022ts2vec} treats enhanced views from the same time step as positive, while views from different time steps as negative. It also introduces instance-wise contrast between samples within the same batch. 
TNC\cite{tonekaboni2021unsupervised} uses a discriminator network to predict subsequent data points.
BTSF\cite{yang2022unsupervised} creates positive pairs by applying a dropout layer to the same sample twice to minimize triplet loss associated with temporal and spectral features.
TF-C\cite{zhang2022self} is proposed to optimize the alignment between the temporal and frequency representations of the same instance.
In addition, CoST\cite{woo2022cost} uses temporal consistencies in the time domain to learn the discriminative trend of a sequence, while transforming the data into the frequency domain to reveal the seasonal representation of the sequence.

\subsection{Knowledge Distillation}
\par The knowledge distillation approach has been successfully applied in several domains, including time series forecasting \cite{hinton2015distilling}. Knowledge distillation is a technique in which a smaller model, called the student model, is trained to mimic the behavior and predictions of a larger and more complex model, called the teacher model. Existing knowledge distillation methods are classified into offline distillation \cite{hinton2015distilling} , online distillation \cite{chen2020online,caron2021emerging}, and self-distillation \cite{mobahi2020self}. 
Based on the type of knowledge, current online knowledge distillation approaches can be mainly categorized into response-based\cite{zhang2018deep}, feature-based\cite{chung2020feature}, and relation-based methods\cite{yang2022mutual}.
Fan et al.\cite{fan2020self} randomly select two subsequences from the same time series and assign pseudo-labels based on their temporal separation. Then, the proposed model is pre-trained to predict the pseudo-labels of the subsequence pairs. Zhang et al.\cite{zhang2021sleeppriorcl} integrate expert features to generate pseudo-labels for self-supervised contrastive representation learning of time series. 
LuPIET\cite{liu2023improving} considers missing data in training to improve predictions with distillation knowledge. However, the prediction of pseudo-labels is prone to include incorrect labels. Therefore, a key focus in the study of training is to mitigate the negative effects of these incorrect labels. 

Recently, some works have exploited the self-distillation framework to achieve good results on time series classification, and other classification and recognition tasks\cite{xiao2023deep,xiao2023capmatch}. In particular, Xiao et.al \cite{xiao2023deep} combines data augmentation, deep contrastive learning, and self-distillation. It considers the contrast similarity of both the high- and low-level semantic information. CapMatch \cite{xiao2023capmatch} focuses on extracting rich representations from input data using a hybrid approach that combines supervised and unsupervised learning techniques. In addition, it construct similarity learning on lower and higher-level semantic information extracted from augmented data to recognize correlations.
Different from existing studies, our work focuses on the integration of representation learning and forecasting into a unified framework. Furthermore, we introduce knowledge distillation between teacher and student models to improve representation performance. We also address the similarity of samples from different timestamps using supervised tasks and incorporate a self-supervised strategy to alleviate data sparsity issues.

\section{Preliminary}\label{3}
\par In this section, we provide essential preliminary concepts and formalize the issue of time series forecasting.
\subsection{Problem Statement} 
Let $X=\{\mathbf {x_1},\mathbf {x_2},...,\mathbf {x_N}\} \in {R}^{N\times T \times C}$ represent a set of time series, where $N$ is the number of  instances. Each vector $\mathbf {x_i}$ has dimension $T \times C$ , where $T$ is the length of time series and $C$ represents the number of channels or features. Given the the look-back window $T$, the time series forecasting problem aims to predict the time series at the next $P$ steps, $\hat{ \mathbf x}=f(\mathbf x)$,
where $\mathbf x\in R^{T\times C}$ is the historical observations, and $\hat{ \mathbf x}\in  R^{P\times C}$ is the future $P$ steps predictions, 
 \begin{figure*}
    \centering
    \includegraphics[scale=0.46]{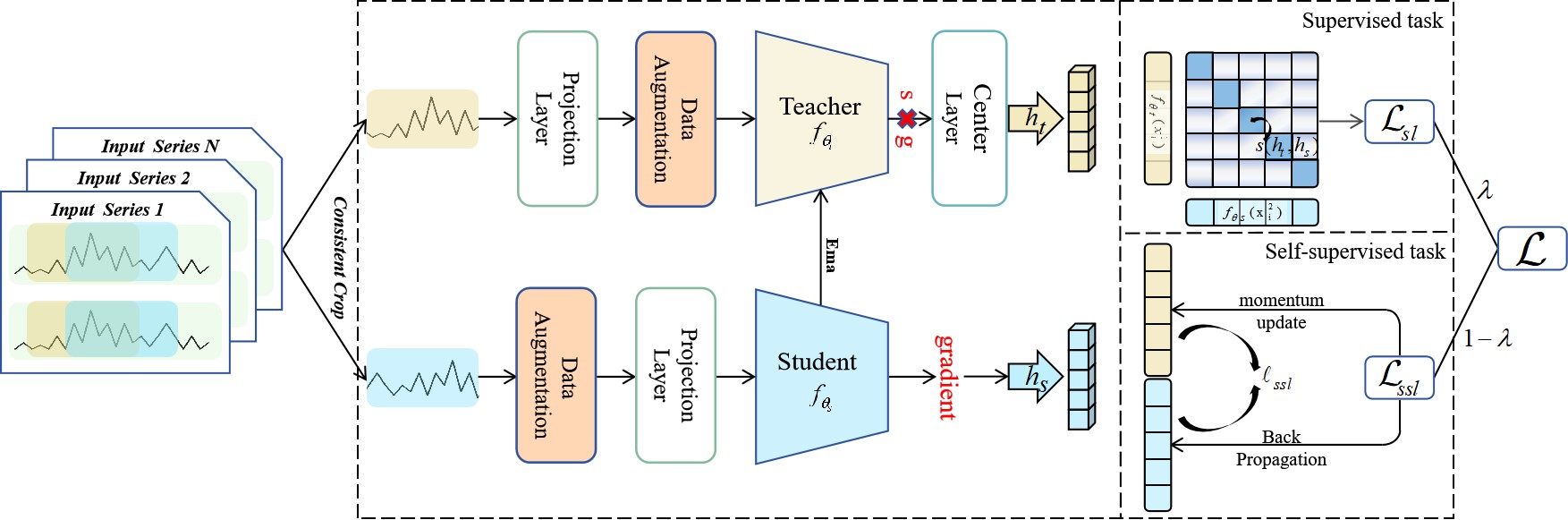}
    \caption{The overall architecture of DE-TSMCL. It consists of four major components: data augmentation, representation learning, supervised task, and self-supervised task.}
    \label{framework}
\end{figure*}
\subsection{Momentum Contrastive Learning}
Consider a sample query $q$ generated by a query encoder and a sequence of samples, denoted $k=\{k_1, k_2, ...,k_N\}$, generated by a momentum encoder. Suppose the unique pairs of $k$ and $q$ in our sample are extremely similar, having been generated by the same encoder with similar parameters. In this case, the objective is to minimize the distance between the most similar pair, represented as $q$ and $k^+$, while increasing the distance between other pairs where the similarity between $q$ and other $k^-$ is not as significant. 

\begin{equation}\label{infoNCE}
\mathcal{L}=-\log\frac{\exp(q,k^+)/\tau}{\exp(q,k^+)/\tau+\sum_{k^-}\exp(q,k^-)/\tau},
\end{equation}

Contrastive learning measures similarity using the dot product. It has been previously referred to as InfoNCE, which is a variation of the NCE loss function initially introduced in CPC\cite{oord2018representation}. Since then, it has been further developed and applied in various contexts such as MOCO\cite{he2020momentum} and GCA\cite{you2020graph}.

\section{METHODOLOGY}\label{4}
In this section, we explain how the DE-TSMCL model works to solve the time series forecasting problem. 
Fig. \ref{framework} shows the overall architecture of our proposed {DE-TSMCL}. In addition, we present a table of notations used in the paper, as depicted in Table 1.

\subsection{Overall Framework}
Our approach follows the teacher-student training scheme. Both the student and teacher networks share the same underlying architecture but possess distinct parameters.
In each training iteration, we first randomly sample the original time series $\mathbf {x}$ to form two overlapping sub-series, namely $\mathbf{x_s}$ and $\mathbf {x_t}$, and perform data augmentation on each sub-series to obtain the augmented view, which are then fed into the teacher and student network, respectively, to generate representation $\mathbf{h_s}$ and $\mathbf {h_t}$. $\mathbf{h_t}$ is utilized as supervision for the teacher network. Next, we set up two tasks: a self-supervise task and an adaptive supervised task. In self-supervised task, we conduct momentum contrastive learning on $\mathbf{h_s}$ and $\mathbf{h_t}$ to capture the complex temporal dependencies among time series and alleviate data noise and incompleteness. In supervised task, we employ a conventional cross-entropy loss to more effectively align the semantic information at the same timestamp, thereby facilitating better updated model parameters through back-propagation. Finally, we jointly optimize the two tasks.

\newcommand{\tabincell}[2]{\begin{tabular}{@{}#1@{}}#2\end{tabular}}

\begin{small}
{
\begin{table}[] 
\label{notation}
\caption{Description of Notations.}
\renewcommand\arraystretch{1.5}
\begin{tabular}{c|l}
\hline \textbf{Notation}     & \multicolumn{1}{c}{\textbf{Description}}  \\
\hline ${X} \in  {R}^{N \times T \times C}$ & \begin{tabular}{l} 
Time series set. $N$ is the number of instances, $T$ is 
the length of look-back window, \\$C$ is the number of 
features.
\end{tabular} \\
 ${H} \in R^{N \times L \times K}$ & \begin{tabular}{l} 
The learned instance representations set. $L$ is the 
length of overlapping subseries,\\ $K$ is the dimension 
of representations.
\end{tabular} \\
 $\mathbf{x}_{\mathbf{i}} \in R^{T \times C}$ &~Time serial of instance $i$. \\
$\mathbf{z}_{\mathbf{i}} \in R^{T \times C}$ & \begin{tabular}{l} 
The latent vector after MLP of instance $i$.
\end{tabular} \\
 $\mathbf{h}_{\mathbf{i}} \in R^{L \times K}$ &~The representation of instance $i$. \\
$B, L$ & \begin{tabular}{l} 
The size of batchsize and overlapping sub-series length.
\end{tabular} \\
$\theta_t, \theta_s$ & \begin{tabular}{l} 
The parameters sets of teacher and student encoders.
\end{tabular} \\
$m, \lambda$ & \begin{tabular}{l} 
The weight for balance loss, and the momentum coefficient.
\end{tabular} \\
$t, i$ &~Timestamp and the order of minibatch. \\
$+,-, \cdot, /$ & \begin{tabular}{l} 
Element-wise addition / minus / multiplication / division.
\end{tabular} \\
\hline
\end{tabular}
\end{table}
}
\end{small}

\subsection{Data Processing and Augmentation}
In order to conserve the macro patterns of time series and facilitate the encoder effectively extract the high-level information, we design the data augmentation block.
The significance of data augmentation in contrastive learning has been acknowledged in previous studies\cite{chen2020simple,grill2020bootstrap}. While existing models commonly employ random masking techniques to refine instance representations, such an approach often introduces biased and noisy information. Moreover, relying solely on masking mechanisms in contrastive learning for time series forecasting is insufficient for generating powerful representations capable of mitigating the aforementioned biases and noises.
To address these limitations, we propose the utilization of parameterized networks for generating optimized representations. Furthermore, the quadratic increase in memory and computational costs associated with the augmentation of multiple views poses practical challenges. To tackle this issue, we introduce a dual-cropping strategy wherein two overlapping sub-series are randomly sampled. In addition, we incorporate learnable augmentation strategies to enhance the robustness of the learned representations.

\subsubsection{Dual-cropping}
We denote the time series $X=\{\mathbf {x_i}\}_{i=1}^{N}$ and the corresponding representations $H=\{\mathbf {h_i}\}_{i=1}^{N}$. 
Firstly, for each instance $\mathbf{x_i}$, we first sample two overlapping sub-series, denoted as $ \mathbf{x_i^1}=\{x_{i,t}\}_{t=a_1}^{b_1}$ and $\mathbf{x_i^2}=\{x_{i,t}\}_{t=a_2}^{b_2}$, where $\{a_1,b_1,a_2,b_2\}\in \{1,2,\cdots, N\}$, satisfying the conditions $a_1<b_1$, $b_1>a_2$, and $a_2<b_2$.
The selection of two sub-sequences from the same time series is crucial as they contain identical semantic information. Without loss of generality, for a given instance $\mathbf {x_i}$, we randomly choose a value $ 0 < L \leq T $ and subsequently extract two overlapping subsequences $\mathbf{x_i^1}$ and $\mathbf{x_i^2}$ of equal window size $T$, and the length of the overlap between two sub-series is $L$.

\subsubsection{Projection Head}
Projection head $p(\cdot)$ maps the original series $\mathbf{x_i}$ into to a high-dimensional latent vector $z_i$. It is meaningful to extract robust, high-quality representations of the input data. For existing methods, multi-layer perception (MLP) with hidden layers is a common choice \cite{wu2019graph,eldele2021time,wang2022revisiting}. In our approach, the projection head consists of a three-layer MLP with $64$ hidden dimensions, followed by $L2$ normalization and a weight-normalized fully connected layer with $K=320$ dimensions. Thus, the latent vector $\mathbf{z_i}$ is represented as,
\begin{equation}
\mathbf{z_i}=p(\mathbf{x_i})= \beta\sigma(\alpha \mathbf{x_i}),
\end{equation}
where $\alpha$ and $\beta$ denote the weights for the hidden layer and output layer, respectively. $\sigma (\cdot)$ is the ReLU activation function. Weights and bias terms are initialized. The initial values of the weight matrix follow the Kaiming initialization strategy, while those of the bias terms follow a uniform distribution. The forward method encapsulates the logic of forward propagation, which involves multiplying the input data by a weight matrix plus a bias term.

\subsubsection{Learnable Data Augmentation}
The purpose of the augmentation is to preserve important features and filter out noisy data, which is the basis for the later selection of positive and negative pairs in contrastive learning. For the teacher network, we mask the high-dimensional mapping $\mathbf{z_i}$ after the MLP. For the student network, we mask the original data $\mathbf{x_i}$ before the MLP. This approach is driven by performance considerations, and we present the detailed analysis in the following experimental section. The two augmentation strategies enhance the representation learning ability of the encoder and make them pay more attention to the valuable information in the time series rather than data noise.
Specifically, we mask the pivotal timestamps of $\mathbf{x_i}$ to generate augmented views, which can be expressed as follows:

\begin{equation}
{g}_{TD}=\{\{x_{i,t}\odot\rho_t\mid x_{i,t}\in\mathcal{X}\},\mathcal{E}\}.
\end{equation}
where $\rho_t\in\{0,1\}$ is drawn from a Bemoulli distribution parameterized by $\omega_t$, i.e., $\rho_t \in Bern(\omega_t)$, which denotes whether to keep the observation $x_{i,t}$ of instance $\mathbf{x_i}$.
Afterwards, the augmented views are respectively fed into the teacher and student networks to obtain the representations.

\subsection{Representation Learning with Knowledge Distillation}
Representation learning aims to learn an encoder $f(X)$, taking the time series as inputs and outputs instance representations in low dimensionality. Let $H=f(X)\in R^{L\times K}$ represent the learned instance representations, where $\mathbf {h_i}$ denote the representation of instance $i$. The generated representations are later employed for prediction. The proposed representation learning is based on knowledge distillation and contrastive learning.

Knowledge distillation is a distinct learning paradigm in which we train a student network $f_{\theta_s}$, to replicate the output of the teacher network $f_{\theta_t}$, $\theta_t$ and $\theta_s$ are the learned parameters of the teacher and student encoder, respectively. Both networks share the same framework while using different parameters set. Through this process, the student network can quickly learn to approximate the teacher network's computational processes for performing the representation task, with less computational overhead in training and inference. 
Specifically, our representation leaning stages is composed of two parts, the encoder and center layer.

 \begin{figure}
    \centering
    \includegraphics[scale=0.261]{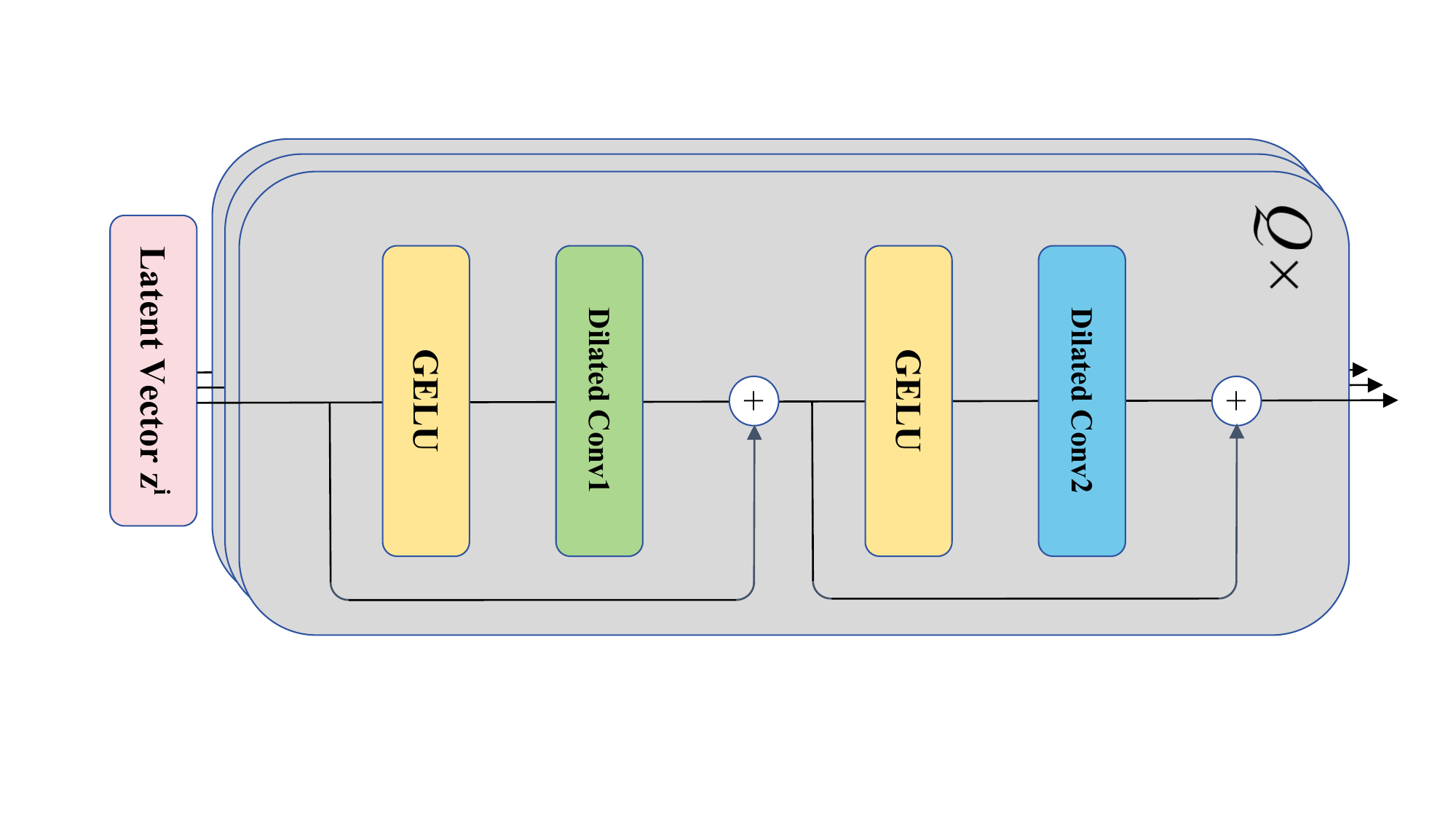}
    \caption{The design of the encoder, where the sequence follows GELU-DilatedConv-GELU-DilatedConv structure.}
    \label{encode}
\end{figure}

\subsubsection{Momentum Encoder}
\label{encoder}
The encoder is designed to extract rich information and generate representations from the augmented time series.
Most existing networks can be applied in our framework, we choose dilated casual convolution and residual connections as the backbone of the encoder. The architecture of the encoder is shown in Fig.\ref{encode}. It stacks multiple layers to extract multiple temporal dependencies in two aspects (i) between different instances, and (ii) between different timestamps of the same instance. The representations after encoding flow in two ways, the contrastive learning representation and the supervised task representation.

We first introduce the dilated causal convolution. Given a 1-D sequence input $\mathbf{z}\in {R}^{n}$ and a function $f:\{0,\ldots,k-1\}\to {R}$, we represent the causal convolution on element $s$ as:
\begin{equation}
    F(s)=(\mathbf{z}*f)(s)=\sum_{i=0}^{k-1}f(i)\cdot\mathbf{z}_{s-i},
\end{equation}

To achieve a larger receptive field for each input, multiple causal convolution layers are usually stacked, so in this paper we use $Q$ as the maximum stacked convolution layers. In most existing works, dilated causal convolution is used to provide the exponential expansion of the receptive field. Formally, the dilated convolution operation $F$ is represented as
\begin{equation}
    F(s)=(\mathbf{z}*_df)(s)=\sum_{i=0}^{k-1}f(i)\cdot\mathbf{z}_{s-d\cdot i},
\end{equation}
where $d$ is the dilation factor, $k$ is the filter size, $d$ increases exponentially according the depth of the network. When $d = 1$, the dilated convolution operator $*d$ reduces to a regular convolution. 

Fig.\ref{encode} shows the backbone of the encoder,  which follows the sequence of GELU $\to $ DilatedConv $\to $ GELU $\to $ DilatedConv. Specifically, each DilatedConv unit comprises two $1-D$ convolutional layers with dilation parameter ($2^p$ is the dilation parameter of $p-$th block, $p\in\{1,2, \cdots, Q\}$).
The Gaussian Error Linear Unit (GELU)\cite{hendrycks2016gaussian} is a powerful activation function defined as $ x\Phi(x) $, where $ \Phi(x) $ is the standard Gaussian cumulative distribution function. GELU is formulated as following,

\begin{equation}
\mathrm{GELU}(x)=x\Phi(x)=x\cdot\frac{1}{2}\left[1+\mathrm{erf}(x/\sqrt{2})\right].
\end{equation}

In addition, residual connections \cite{he2016deep} between adjacent layers are employed to facilitate the model ability to acquire deeper semantic information. These components cooperate to jointly extract the context representation for each timestamp. 

In addition to the TCN\cite{bai2018empirical} used in our encoder, we can incorporate other convolutional frameworks into our model, such as sparse convolutions \cite{buza2023sparsity, tian2023designing}, which have shown promise in the field of 3D reconstruction. Notably, the study presented in \cite{buza2023sparsity} introduces the concept of sparsity-invariant convolution for handling irregularly sampled time series. This research offers valuable insights that can significantly contribute to solving the challenge of forecasting irregularly sampled time series using our proposed model.

\subsubsection{Center Layer}
In order to enhance the robustness and stability of the model, achieve better representation learning and more reliable performance in downstream tasks, inspired by DINO\cite{caron2021emerging}, we employ a center layer to mitigate the impact of distance, promote uniform convergence, and prevent the dominance of dimensions.

In particular, by adding a centering term to the output of the function $f_{\theta_t}(x)$, the center layer reduces the influence of the distances between diverse samples on the loss calculation. It is formulated as:
\begin{equation}
\begin{aligned}
f_{\theta_t}(x)\leftarrow f_{\theta_t}(x)+c,
\end{aligned}
\end{equation}
where the mean operator is implemented along the temporal dimension, represented as:
\begin{equation}
\begin{aligned}
c=\frac{1}{B}\sum_{i=1}^{B}f_{\theta_t}(x_{i}).
\end{aligned}
\end{equation}

The advantages of centering operation are as following three-folds: (1)
It increases the robustness of the model by reducing the sensitivity to variations in distances between different samples. This robustness allows the model to better handle variations in data distribution and improves its generalization capabilities. (2) It promotes stable training dynamics and ensures that the model captures meaningful representations. (3)
It enhances the discriminative ability of the learned representations and leads to more effective separation of classes or clusters in the learned embedding space, thereby enhancing the performance of downstream tasks.

\subsection{Momentum Contrastive Learning Task}
In this section, we introduce the designed self-supervised learning task. A good representation learning mapping is capable of maximizing the similarity between correlated instances in the representation space.
After obtaining the representations from the teacher and student networks, the selection of appropriate positive and negative samples is crucial for contrastive learning. Positive pairs are used to emphasize the consistency between different views of the same data point.
Negative pairs, on the other hand, are used to enforce separation between different data points, by encouraging the model to learn distinct features for each data point.
Once the positive and negative samples have been identified, they can then be used to train the model to learn more robust feature representations, which can then be applied to a range of downstream tasks.

In this article, inspired by \cite{radford2021learning}, in-batch sampling strategy is adopted to construct positive and negative pairs. Given a bath of paired representation $(h_{i,t_1}^{t}, h_{j,t_2}^{s})$, $i,j\in\{1,2,\cdots,B\}$ is the batch number, and $t_1,t_2\in\{1,2,\cdots,L\}$ is the timestamp. When $i=j$ it is a positive pair. For negative pairs, we choose from the following two aspects:
\begin{itemize}
    \item Temporal view: $i\neq j$ and $t_1=t_2$, $(h_{i,t_1}^{t}, h_{j,t_2}^{s})$ is negative pair.
    \item Spatial view: $i=j$ and $t_1 \neq t_2$, $(h_{i,t_1}^{t}, h_{j,t_2}^{s})$ is negative pair.
\end{itemize}

Having chosen the positive and negative samples, we utilize the classical InfoNCE loss \cite{you2020graph,zhu2021graph,he2020momentum}, to maximize the similarity between positive pairs, while minimizing the similarity between negative pairs. The self-supervised contrastive loss for the $i$-th instance at timestamp $t$ can be formulated as follows:

    \begin{equation}
\ell_{ssl}^{(i,t)}=\log\frac{-(\exp(h_{i,t}^{t}\cdot h_{i,t}^{s}))}{\sum\limits_{j=1}^{B}\sum\limits_{t'^{^{}}=1}^{L}\left(\exp\left.(h_{i,t}^{t}\cdot\left.(h_{i,t^{\prime}}^{s}+h_{j,t}^{s})\right.)\right.+\exp\left.(h_{i,t}^{s}\cdot(\left.h_{j,t}^{s}+h_{i,t^{\prime}}^{s})\right.)\right.\right)}
    \end{equation} 
where $t, t^{\prime} \in \{1,2,\cdots,L\}$, $ t \neq t^{\prime} $. $i,j \in \{1,2,\cdots,B\}$, $i\neq j$.
Finally, the contrastive learning loss over the entire time series is expressed as:
\begin{equation}
\begin{aligned}
\mathcal{L}_{ssl}=\frac{1}{NL}\sum_{i}\sum_{t}\left(\ell_{ssl}^{(i,t)}\right),
\end{aligned}
\end{equation}

In real-world applications, the inherent variability of time series data can indeed pose challenges in the training of models. Short-term errors or fluctuations can have a significant impact on the learning process, leading to the model focusing on abnormal or noisy features. Momentum contrastive learning is one approach that can help address the issue of unstable or noisy negative pairs in contrastive learning, resulting in more effective representation learning and improved performance on downstream tasks\cite{he2020momentum}.
Instead of comparing positive and negative pairs directly, it uses a momentum encoder to create a moving-average of the encoder weights over multiple steps. 
With its ability to consider historical observations and provide more stable negative pairs for contrastive learning, momentum contrastive learning can further improve the accuracy of predictions by incorporating valuable contextual information from past timestamps. By leveraging this approach, we can potentially enhance the modeling and understanding of the underlying temporal dynamics in data. Meanwhile, more stable and informative comparisons are achieved.

Given the parameters sets $\theta_t$ and $\theta_s$ of $f_{\theta_t}$ and $f_{\theta_s}$, the gradient for $f_{\theta_t}$ is updated as follows:
\begin{equation}\label{m-update}
   \theta_t = m\theta_t + (1 - m)\theta_s.
\end{equation}
where momentum coefficient $m\in [0,1)$. As shown in Fig.\ref{framework}, back-propagation is employed to update $\theta_s$, while Eq.(\ref{m-update}) is used to update $\theta_t$.
Inspired by the work of DINO\cite{caron2021emerging}, we utilize the stop gradient operator (s-g) on the teacher network to restrict the propagation of gradients exclusively through the student network. This ensures that during the back-propagation process, the gradients are not updated for the teacher network. Instead, the teacher parameters are updated using the exponential moving average (Ema) of the student parameters. Conversely, for the student network, the gradients are updated normally, allowing for the learning and optimization of its parameters. To maintain consistency and visual clarity in representing the flow of gradients within the teacher-student network, we also include a gradient return marker of the same color for the student network. This decision aligns with our pseudocode and promotes a uniform representation of operators throughout the network architecture.

\par 
As discussed MOCO\cite{he2020momentum}, the model shows superior performance in image classification when $m=0.999$. In our experiments, we explore the influence of $m$ on the model performance, and achieve the best result when $m=0.999$.
By incorporating a slowly updating encoder parameter, the model can adapt more gradually to changes in the data distribution over time, which allows the model to better capture the long-term dependencies and patterns in the data, while also providing a certain level of robustness against distribution shifts.

\subsection{Adaptive Supervised Task}
In this section, we design a supervised learning task as the auxiliary task to learn more robust representations and facilitate the contrastive learning process. This multi-task learning framework leverages the shared representations and relationships among different tasks to enhance the overall performance of the model. Using the teacher's soft labels to teach the student mitigates the effects of mislabeling that result from the absence of the teacher-student mechanism in previous research. In particular, we utilize the output of the teacher network as the soft label supervision signal for the student network. Soft labels provide a means of transferring knowledge from teacher network to a student network model during distillation.
The teacher in DE-TSMCL generates labels that are semantically correct, structurally similar, and smoother compared to the potentially noisy or incorrect labels. This is achieved through knowledge distillation technique, which is represented as:

\begin{equation}
\begin{aligned}
p_i^t = softmax(\mathbf{h^t_{i}}) = \frac{\exp(\mathbf{h_{i}^t})}{\sum_{j=1}^{L}\exp(\mathbf{h_{j}^t})},
\\
p_i^s =softmax(\mathbf{h^s_{i}}) = \frac{\exp(\mathbf{h_{i}^s})}{\sum_{j=1}^{L}\exp(\mathbf{h_{j}^s})},
\end{aligned}
\end{equation}
where $\mathbf{h^t_{i}}$ and $\mathbf{h^s_{i}}$ are the representations of instance $i$ from the teacher and student model, respectively. The soft labels from the teacher provide a more reliable and informative training signal, helping the student model to focus on the meaningful patterns and learn more accurate and robust representations from the training data.

Finally, we employ the prevalent cross-entropy loss\cite{fan2020self} as the supervised loss function to control the optimization of the module. The objective of the cross-entropy loss is to compute the similarity $s(h_t,h_s)$ between two probability matrices, the two probabilities $p_i^s$ and $p_i^t$ after softmax correspond to the two probability distributions in the cross-entropy loss, which is equivalent to implementing MLE (Maximum Likelihood Estimation) on $p_i^s$ and $p_i^t$, i.e., making them more similar, consistent with our conceptualization of evaluating not only the distances between positive and negative
samples within the sample space, but also the comparability of their respective pairs. It is formulated as:

\begin{equation}
\mathcal L_{sl} = \sum_{i=1}^L s(h_t,h_s) =-\sum_{i=1}^L p_i^s \log ( p_i^t ).
\end{equation}

By minimizing the cross-entropy loss, the model learns to make more accurate predictions and better differentiate between different instances. 

\definecolor{codegreen}{rgb}{0,0.6,0}
\definecolor{codegray}{rgb}{0.5,0.5,0.5}
\definecolor{codepurple}{rgb}{0.58,0,0.82}

\lstdefinestyle{mystyle}{
  commentstyle=\color{codegreen},
  keywordstyle=\color{blue},
  numberstyle=\tiny\color{codegray},
  stringstyle=\color{codepurple},
  basicstyle=\ttfamily\footnotesize,
  breakatwhitespace=false,         
  breaklines=true,                  
  captionpos=b,                    
  keepspaces=true,                  
  numbers=left,                    
  numbersep=5pt,                  
  showspaces=false,                
  showstringspaces=false,
  showtabs=false,                  
  tabsize=2
}

\begin{algorithm}
\caption{DE-TSMCL Pre-train PyTorch-like style pseudocode.}
\label{alg:mymoco}
\begin{algorithmic}
\item \lstset{style=mystyle}
\begin{lstlisting}[language=Python]
# f_t,f_s: encoder networks for teacher & student
# mask : binomial mask  # crop : random dual-crop
# lambda : weight parameter
# m : momentum coefficient
# c : center()

f_t.params = f_s.params # initialize
for x in loader: # load minibatch x with N samples
    x_t = crop(x) # consistent crop
    x_s = mask( crop(x) ) # data aug
    
    z_t,z_s = projection(x_t,x_s) # MLP

    # teacher encoder
    h_t = f_t.forward( mask(z_t) )
    # student encoder
    h_s = f_s.forward(z_s)

    loss = combine(h_t,h_s) # joint optimization
    loss.backward()

    # student updates
    update(f_s)  # SGD update(just student)
    # teacher updates
    f_t.params = m*f_t.params + (1-m) * f_s.params
def combine(t,s) # momentum updates
    t = t.detach() # stop-gradient
    p_t = softmax(t-c) # softlabel with center
    p_s = softmax(s) # predlabel
    sl = EntrpyLoss(p_t,p_s) # similarity
    ssl = InfoNCE(s,t) # contrastive learning
    return lambda * sl + (1-lambda) * sll
\end{lstlisting}
\end{algorithmic}
\end{algorithm}

\subsection{Joint Optimization and Prediction}
The loss function for DE-TSMCL consists of two loss terms: a self-supervised loss $\mathcal L_{\text{ssl}}$ and a supervised loss $\mathcal L_{\text{sl}}$. As the sampled two overlapping sub-series have certain overlapping segments, which prompts a focus not only on positive and negative pair distances within the sample space during contrastive learning, but also on the acquisition of similar semantic information between the overlapping segments. 

Therefore, in order to combine the above two tasks, we jointly optimize the model loss as follows,
 \begin{equation}\label{joint}
\mathcal{L}=\lambda\mathcal L_{\text{sl}}+\left(1-\lambda\right)\mathcal L_{\mathrm{ssl}},
 \end{equation}
where $\lambda$ is a hyperparameter representing the relative weight of the supervised task.
The experimental results have shown that this joint optimization is capable of improving the model's representation learning capabilities and boosting its performance in downstream tasks.

The final step of DE-TSMCL is prediction. When the pre-training procedure has been accomplished, we integrate the pre-trained encoder $f_{\theta_s}$ with the following prediction module. We feed the learned representation into a {Linear} layer to generate predicted result with the same dimensionality as the ground truth {$y_i$}, 

\begin{equation}
  \hat {\mathbf{y_i}}=\text{Linear}(\mathbf{h_i^s})=\text{Linear}((f_{\theta_s}(\mathbf{z}_{i}))),
\end{equation}
where $\hat {\mathbf{y_i}}$ is the predicted value, $\mathbf{y_i}$ is the ground truth. Then, we train the model using Mean Absolute Error (MAE) as the regression loss.
Ridge regression is a regularization technique commonly employed in regression tasks, including time series forecasting. It introduces a penalty term to the loss function, which helps mitigate overfitting and stabilize the model's predictions. In our model, we use cross-validation to identify the optimal ridge regression model for the forecasting task, which is then used in subsequent forecasting stage.

\begin{equation}
   \begin{aligned}J(
   \theta) & =\frac1m\sum_{i=1}^{m}\left.\left(\theta^{T}\cdot y_i-x_i\right)^2+\alpha\sum_{i=1}^{n}\theta_{i}^2\right..
   \end{aligned}
\end{equation}
 Let $\theta$ denote all parameters, and $\alpha$ denote a hyperparameter that balances the relative importance of the norm penalty term $\left.\omega=\sum_{i=1}^{n}\theta_{i}^2{}\right.$ and the standard objective function $J(\theta)$.

An overview of the training process for DE-TSMCL is provided in Algorithm 1. DE-TSMCL is an efficient algorithm when compared to traditional contrastive learning models. The complexity of DE-TSMCL is $O(TCK) + O(T) + O(KL) + O(KL^2)$, where $O(TCK)$ denotes the complexity of the projection layer, $O(T)$ is the complexity of the data augmentation module, the complexity of the momentum encoder and supervised loss is $O(KL)$, and $O(KL^2)$ represents the complexity of the InfoNCE loss. In DE-TSMCL, the sub-series length $T$ scales linearly with both the projection layer and augmentations. The joint optimization module scales quadratically with the number of overlapping sub-series $L$ and linearly with the dimension $K$.


\begin{table*}[]
\centering
\caption{Summary of the popular datasets for benchmark.}
\scalebox{1.0}{
\begin{tabular}{lcccccccc}
\toprule
Datasets      & \multicolumn{1}{l}{Channels} & \multicolumn{1}{l}{Granularity} & \multicolumn{1}{l}{Length} & \multicolumn{1}{l}{Training} & \multicolumn{1}{l}{Validation} & \multicolumn{1}{l}{Testing} \\ 
\midrule
ETTh1  & 7   & 1 hour   & 17420  &60\%  &20\% &20\% \\  
ETTh2   & 7   & 1 hour   & 17420      &60\%  &20\% &20\% \\ 
ETTm1      & 7   & 15 minutes   & 69680   &60\%  &20\% &20\%  \\ 
ETTm2      & 7   & 15 minutes   & 69680   &60\%  &20\% &20\%  \\ 
\midrule
Electricity       & 321   & 1 hour   & 26304   &60\%  &20\% &20\% \\
\bottomrule
\end{tabular}
}\label{dataset}
\end{table*}

\section{Experimental Results and Analysis}\label{5}
In this section, we first compare the proposed solution with state-of-the-art methods, and then validate the effectiveness of each component through extensive ablation studies. In particular, we investigate the following research questions:
\begin{itemize}
    \item \textbf{RQ1.} Does the proposed {DE-TSMCL} outperform existing baseline methods on time series prediction problem?
    \item \textbf{RQ2.} Do all modules of the model contribute to the overall performance of {DE-TSMCL}? How does each module impact the model performance?
    \item \textbf{RQ3.} How do the proposed momentum update and centering techniques contribute to the encoder? How does the joint optimization impact the model performance?
     \item \textbf{RQ4.} How does the learned data augmentation influence the the model performance?
\end{itemize}
For comprehensively evaluating the model, we study the performance of {DE-TSMCL} for solving univariate and multivariate time series forecasting tasks, respectively.

\subsection{Experimental Settings}
\subsubsection{Dataset Description}
We conduct experiments on five public real-world datasets, including ETTh1, ETTh2, ETTm1, ETTm2 and Electricity. The summarized statistics of the datasets are presented in Table \ref{dataset}. Following existing works\cite{yue2022ts2vec,woo2022cost}, we partition the data into training, validation testing set with a ratio of 3:1:1 in the experiments.
\begin{itemize}
    \item \textbf{ETT}(Electricity Transformer Temperature)\cite{zhou2021informer}: 
    The ETT serves as a key indicator in the long-term deployment of electric power. This dataset encompasses data spanning two years from two distinct counties in China. In order to delve into the granularity of the long sequence time-series forecasting issue, various subsets are formulated: {ETTh1, ETTh2} for the 1-hour level, and ETTm1,ETTm2 for the 15-minute level. Each data point comprises the target value "oil temperature" along with six power load features. 
    \item \textbf{Electricity}\cite{wu2021autoformer}: The electricity dataset is the electric power consumption measurements are taken in one household at a one-minute sampling rate over a nearly 4-year period. Various electrical quantities and some sub-metering values are available. This dataset contains 2,075,259 measurements collected in a house located in Sceaux, France (7 km from Paris) between December 2006 and November 2010 (spanning 47 months).
\end{itemize}


\subsubsection{\textbf{Compared Methods}}
We compare DE-TSMCL with nine state-of-the-art baseline methods. These baselines can be divided into two categories: End-to-end models and Representation learning models. 
\par The End-to-end Forecasting models include:
\begin{itemize}  
    \item \textbf{LSTNet} \cite{lai2018modeling}:  Using LSTM networks for time series forecasting. This variant demonstrates a superior capacity to encapsulate long-term correlations within the time series.
    \item  \textbf{TCN} \cite{bai2018empirical}: A time series forecasting model that uses CNN for sequential data processing. The convolution operation is employed for processing time series data, rather than using a loop operation. It extends the receptive field by implementing multiple convolutional blocks to capture long-distance dependencies.
    \item  \textbf{N-BEATS} \cite{oreshkin2019n}: It is a block-based network structure to learn intricate timing dependencies. The prediction results from various modules are integrated through combination operations to accomplish multi-view time series modeling.
    \item \textbf{LogTrans} \cite{li2019enhancing}: It utilizes log-transformed series data for improved time series predictions. Address the issue of non-static trends present in the original data.
    \item \textbf{Reformer} \cite{kitaev2020reformer}: It introduces local-sensitive hashing (LSH) to approximate attention by allocating similar queries. Not only are subsequent models improved by reduced complexity, but they also further develop intricate building blocks for time series forecasting.
    \item \textbf{Autoformer} \cite{wu2021autoformer}: It utilizes autocorrelation to establish patch-level connections, but this is a manually designed approach that does not encapsulate all semantic information within a patch.
    \item \textbf{Informer} \cite{zhou2021informer}: It presents an enhanced transformer architecture for time series forecasting.  It mitigates the complexity of self-attention by employing random sampling and probability distribution.
\end{itemize}

The representation learning models include:
\begin{itemize}
    \item \textbf{CPC} \cite{oord2018representation}: It learns representations by predicting future samples in latent space using auto-regressive models. It employs a probabilistic contrastive loss, which encourages the latent space to maximize the capture of information useful for predicting future samples.
    \item \textbf{MoCo} \cite{he2020momentum} : An unsupervised learning method, mainly applied in visual representation learning in deep learning, learns data characteristics through momentum comparison.
    \item \textbf{TNC} \cite{tonekaboni2021unsupervised}: Time series prediction employing convolutional neural networks and Fourier transformation in both time and frequency domains. Integrating time-frequency multipurpose features allows for a comprehensive identification of time series outliers from various perspectives.
    \item \textbf{TS2Vec}  \cite{yue2022ts2vec}: An embedding learning approach that maps time-series data into vectors. Time series data is conceptualized as a spatiotemporal graph, and the representation learning technique from image processing is employed. By implementing multi-level convolution and pooling operations, the time series is processed into a fixed-length vector.
\end{itemize}

\subsubsection{\textbf{Evaluation Metrics}}
We evaluate different approaches with two representative evaluation metrics in the field of time series prediction: Mean Squared Error (MSE) and Mean Absolute Error (MAE)\cite{zhou2021informer,wu2021autoformer,yue2022ts2vec}. The metrics are defined as follows:

\begin{equation}
\begin{aligned}
\mathrm{MSE}=\frac{1}{PC}\sum_{i=1}^{P}\sum_{j=1}^{C}\left(x_{t+i}^{(j)}-\hat{x}_{t+i}^{(j)}\right)^2,
\end{aligned}
\end{equation}

\begin{equation}
\begin{aligned}
\mathrm{MAE}=\frac{1}{PC}\sum_{i=1}^{P}\sum_{j=1}^{C}|x_{t+i}^{(j)}-\hat{x}_{t+i}^{(j)}|.
\end{aligned}
\end{equation}
where $\hat{x}_{t+i}^{(j)}$ and $x_{t+i}^{(j)}$ are the predicted value and the ground truth of instance $j$ at time step $t+i$.

\subsubsection{\textbf{Training Details $\&$ Hyperparameters}}
The model has been implemented based on the PyTorch framework. All experiments have been carried out on a Intel(R) Core i5-8400 $@$ 2.80GHz hardware platform equipped with NVIDIA GeForce GTX 1060(6G) GPU.  
Specially, the default batch size is 4, and the learning rate is $1e^{-3}$. For ETTh1 and ETTh2 datasets, the number of pre-training iterations is 200, while for ETTm1, ETTm2 and electricity datasets, it is increased to 600. The dimension of representations is 320. The channels of MLP projection head  hidden layers is 64.
Following existing works such as TS2vec\cite{yue2022ts2vec} and Informer\cite{zhou2021informer}, the prediction length is $P$ $ \in $ \{24, 48, 96, 288, 672\} for the ETTm1 dataset and $P$ $ \in $ \{24, 48, 168, 336, 720\} for the remaining datasets.

\par Similar to Ts2vec\cite{yue2022ts2vec}, a linear regression model with $L2$ regularization $\alpha$ is trained on the learned representations to predict the future values. $L2$ regularization allows the learning algorithm to "perceive" input $x$ with higher variance, resulting in a shrinking of the weights associated with less informative features. Consequently, using $L2$ regularization allows less informative features to be neglected, thereby preventing over-fitting. We utilize the validation set to determine the optimal ridge regression regularization term $\alpha$ within a search space consisting of \{0.1, 0.2, 0.5, 1, 2, 5, 10, 20, 50, 100, 200, 500, 1000\}.

\begin{table*}[]
\centering
\caption{Comparison with baselines for univariate time series forecasting. \textbf{Bold} represents the best performance.}
\label{comparison_uni}
\scalebox{0.55}
{
\begin{tabular}{ll|ccccc|ccccc|ccccc|ccccc}
\toprule
\multicolumn{2}{l|}{Datasets}        & \multicolumn{5}{c|}{ETTh1} & \multicolumn{5}{c|}{ETTh2} & \multicolumn{5}{c|}{ETTm1} & \multicolumn{5}{c}{Electricity} \\
\midrule
Methods                   & Metrics &24 &48 &168 &336 &720  &24 &48 &168 &336 &720   &24 &48 &96 &288 &672  &24     &48     &168     &336  &720  \\
\midrule
\multirow{2}{*}{{LSTnet}}       &$MSE(\downarrow)$      
&0.108 &0.175 &0.396 &0.468 &659   &3.554 &3.190 &2.800 &2.753 &2.878
&0.090 &0.179 &0.272 &0.462 &0.639    &0.281 &0.381 &0.599 &0.823 &1.278\\
                          & $MAE(\downarrow)$   
                     &0.284 &0.424 &0.504 &0.593 &0.776 
 &0.445 &0.474 &0.595 &0.738 &1.044 
&0.206 &0.306 &0.399 &0.558  &0.697
 &0.287 &0.366 &0.500 &0.624  &0.906 \\
\midrule 
\multirow{2}{*}{{TCN}}       &$MSE(\downarrow)$ 
&0.075 &0.227 &0.316 &0.306 &0.390  &0.075 &0.227 &0.316 &0.306 &0.325
&0.041 &0.101 &0.142 &0.318 &0.397   &0.263 &0.373 &0.609 &0.855 &1.263\\
                          &$MAE(\downarrow)$
&0.210 &0.402 &0.493 &0.495 &0.557     &0.249 &0.290 &0.376 &0.430 &0.463 
&0.157 &0.257 &0.311 &0.472 &0.547     &0.279 &0.344 &0.462 &0.606 &0.858\\
\midrule 
\multirow{2}{*}{{N-BEATS}}     & $MSE(\downarrow)$ 
&0.094 &0.210 &0.232 &0.232 &0.322     &  0.198  &0.234  &0.331 &0.431 &0.437      
&0.054 &0.190 &0.183 &0.186 &0.197      & 0.427 &0.551 &0.893 &1.035 &1.548  \\
                          & $MAE(\downarrow)$  
&0.238 &0.367 &0.391 &0.388 &0.490  &0.345 &0.386 &0.453 &0.508  &0.517      
&0.184 &0.361 &0.353 &0.362 &0.368     &0.330 &0.392 &0.538 &0.669  &0.881 \\
\midrule                          
\multirow{2}{*}{{LogTrans}}&$MSE(\downarrow)$
&0.103 &0.167 &0.207 &0.230 &0.279  & 0.102 &0.169 &0.246 &0.267 &0.303    
&0.065 &0.078 &0.199 &0.411 &0.598    &0.528 &0.409 &0.959 &1.079 &1.001\\  
                          & $MAE(\downarrow)$
&0.259 &0.328 &0.375 &0.398 &0.463    &0.255 &0.348 &0.422 &0.437 &0.493  
& 0.202 &0.220 &0.386 &0.572 &0.702   &0.447 &0.414 &0.612 &0.639 &0.714 \\
\midrule                          
\multirow{2}{*}{{CPC}} &$MSE(\downarrow)$   
&0.076 &0.104 &0.162 &0.183 &0.212    &0.109 &0.152 &0.251 &0.238 &0.234 
&{0.018} &0.035 &0.059 &0.118 &0.177  &0.264 &0.321 &0.438 &0.599 &0.957 \\
                          & $MAE(\downarrow)$ 
&0.217 &0.259 &0.326 &0.351 &0.387   &0.251 &0.301 &0.392 &0.388 &0.389   
&0.102 &0.142 &0.188 &0.271 &0.332   &0.299 &0.339 &0.418 &0.507 &0.679 \\
\midrule                          
\multirow{2}{*}{{MoCo}} &$MSE(\downarrow)$    
&0.040 &0.063 &0.122 &0.144 &0.183    &0.095 &0.130 &0.204 &0.206 &0.206 
&{0.015} &0.027 &0.041 &0.083 &0.122  &0.254 &0.304 &0.416 &0.556 &0.858 \\
                          & $MAE(\downarrow)$ 
&0.151 &0.191 &0.268 &0.297 &0.347   &0.234 &0.279 &0.360 &0.364 &0.369   
&0.091 &0.122 &0.153 &0.219 &0.268   &0.280 &0.314 &0.391 &0.482 &0.653
  \\
\midrule                          
\multirow{2}{*}{{TNC}} &$MSE(\downarrow)$ 
&0.057 &0.094 &0.171 &0.192 &0.235   &0.097 &0.131 &{0.197} &0.207 &0.207  
&0.019 &0.036 &0.054 &0.098 &0.136    &0.252 &0.300 &0.412 &0.548 &0.859
\\
 & $MAE(\downarrow)$ 
&0.184 &0.239 &0.329 &0.357 &0.408
&0.238 &0.281 &0.354 &0.366 &0.370
&0.103 &0.142 &0.178 &0.244 &0.290
&0.278 &0.308 &0.384 &0.466 &0.651
\\
\midrule
\multirow{2}{*}{{Informer}}     &$MSE(\downarrow)$ 
&0.098 &0.158 &0.183 &0.222 &0.269    &0.093 &0.155 &0.232 &0.263 &0.277 
&0.030 &0.069 &0.194 &0.401 &0.512    &0.251 &0.346 &0.544 &0.713 &1.182\\
&$MAE(\downarrow)$ 
&0.247 &0.319 &0.346 &0.387 &0.435   &0.240 &0.314 &0.389 &0.417 &0.431
&0.137 &0.203 &0.372 &0.554 &0.644   &0.275 &0.339 &0.424 &0.512 &0.806 \\
\midrule                          
\multirow{2}{*}{{TS2Vec}} &$MSE(\downarrow)$
&0.039 &0.062 &0.134 &0.154 &0.163
&0.091 &0.124 &0.208 &0.213 &0.214
&0.015 &0.027 &0.044 &0.103 &0.156
&0.260 &0.319 &0.427 &0.565 &0.861 \\
& $MAE(\downarrow)$  
&0.152 &0.191 &0.282 &0.310 &0.327
&0.229 &0.273 &0.360 &{0.369}&0.374
&0.092 &0.126 &0.161 &0.246 &0.307
&0.288 &0.324 &0.394 &0.474 &0.643\\  
\midrule  
\midrule

\multirow{2}{*}{\textbf{DE-TSMCL}}     &$MSE(\downarrow)$ 
&\textbf{0.038} &\textbf{0.059} &\textbf{0.115} &\textbf{0.135} &\textbf{0.157}  
&\textbf{0.088} &\textbf{0.120} &\textbf{0.191} &\textbf{0.198} &\textbf{0.200}        
&\textbf{0.013} &\textbf{0.022} &\textbf{0.036} &\textbf{0.079}  &\textbf{0.119}    
&\textbf{0.248} &\textbf{0.294} &\textbf{0.403} &\textbf{0.537} &\textbf{0.843}  \\
& $MAE(\downarrow)$ 
&\textbf{0.146} &\textbf{0.181} &\textbf{0.260} &\textbf{0.285} &\textbf{0.310}
&\textbf{0.229} &\textbf{0.272} &\textbf{0.351} &\textbf{0.359}  
&\textbf{0.363}  
&\textbf{0.083} &\textbf{0.110} &\textbf{0.143} &\textbf{0.211}  &\textbf{0.262}  
&\textbf{0.272} &\textbf{0.301} &\textbf{0.373} &\textbf{0.458}  &\textbf{0.644}  \\
\bottomrule
\end{tabular}}
\end{table*}

\begin{table*}[]
\centering
\caption{Comparison with baselines for multivariate time series forecasting. \textbf{Bold} represents the best performance.}
\label{comparison_mu}
\scalebox{0.55}{
\begin{tabular}{ll|ccccc|ccccc|ccccc|ccccc}
\toprule
\multicolumn{2}{l|}{Datasets}        & \multicolumn{5}{c|}{ETTh1} & \multicolumn{5}{c|}{ETTh2} & \multicolumn{5}{c|}{ETTm1} & \multicolumn{5}{c}{Electricity} \\
\midrule
Methods                   & Metrics &24 &48 &168 &336 &720  &24 &48 &168 &336 &720   &24 &48 &96 &288 &672  &24     &48     &168     &336  &720  \\
\midrule
\multirow{2}{*}{{LSTnet}}       &$MSE(\downarrow)$      
&1.293 &1.456 &1.997 &2.655 &2.143  &2.742 &3.567 &3.242 &2.544 &4.625
&1.968 &1.999 &2.762 &1.257 &1.917   &0.356 &0.429 &0.372 &0.352 &0.380\\
                      & $MAE(\downarrow)$   
                     &0.901 &0.960 &1.214 &1.369 &1.380
 &1.457 &1.687 &2.513 &2.591 &3.709
 &1.170 &1.215 &1.542 &2.076 &2.941
 &0.419 &0.456 &0.425 &0.409 &0.443 \\
\midrule 
\multirow{2}{*}{{TCN}}       &$MSE(\downarrow)$ 
&0.767 &0.713 &0.995 &1.175 &1.453  &1.365 &1.395 &3.166 &3.256 &3.690
&0.324 &0.477 &0.636 &1.270 &1.381  &0.305 &0.317 &0.358 &0.349 &0.447\\
&$MAE(\downarrow)$
&0.612 &0.617 &0.738 &0.800 &1.311     &0.888 &0.960 &1.407 &1.481 &1.588
&0.374 &0.450 &0.602 &1.351 &1.467     &0.384 &0.392 &0.423 &0.416 &0.486\\
\midrule                          
\multirow{2}{*}{{LogTrans}}&$MSE(\downarrow)$
&0.686 &0.766 &1.002 &1.362 &1.397  & 0.828 &1.806 &4.070 &3.875 &3.913    
&0.419 &0.507 &0.768 &1.462 &1.669    &0.297 &0.316 &0.426 &0.365 &0.344\\  
& $MAE(\downarrow)$
&0.604 &0.757 &0.846 &0.952 &1.291    &0.750 &1.034 &1.681 &1.763 &1.552  
&0.412 &0.583 &0.792 &1.320 &1.461  &0.374 &0.389 &0.466 &0.417 &0.403 \\
\midrule                          
\multirow{2}{*}{{CPC}} &$MSE(\downarrow)$   
&0.728 &0.774 &0.920 &1.050 &1.160    &0.551 &0.752 &2.452 &2.664 &2.863
&0.478 &0.641 &0.707 &0.781 &0.880   &0.403 &0.424 &0.450 &0.466 &0.559 \\
& $MAE(\downarrow)$ 
&0.600 &0.629 &0.714 &0.779 &0.835   &0.572 &0.684 &1.213 &1.304 &1.399  
&0.459 &0.550 &0.593 &0.644 &0.700   &0.459 &0.473 &0.491 &0.501 &0.555 \\
\midrule                          
\multirow{2}{*}{{MoCo}} &$MSE(\downarrow)$    
&0.623 &0.669 &0.820 &0.981 &1.138    &0.444 &0.613 &1.791 &2.241 &2.425
&0.458 &0.594 &0.621 &0.700 &0.821    &0.288 &0.310 &0.337 &0.353 &0.380 \\
                          & $MAE(\downarrow)$ 
&0.555 &0.586 &0.674 &0.755 &0.831    &0.495 &0.595 &1.034 &1.186 &1.292 
&0.444 &0.528 &0.553 &0.606 &0.674   &0.374 &0.390 &0.410 &0.422 &0.441
  \\
\midrule                          
\multirow{2}{*}{{TNC}} &$MSE(\downarrow)$ 
&0.708 &0.749 &0.884 &1.020 &1.157     &0.612 &0.840 &2.359 &2.782 &2.753
&0.522 &0.695 &0.731 &0.818 &0.932     &0.354  &0.376  &0.402 &0.417 &0.442\\
 & $MAE(\downarrow)$ 
&0.592 &0.619 &0.699 &0.768 &0.830    &0.595 &0.716 &1.213 &1.349 &1.394
&0.472 &0.567 &0.595 &0.649 &0.712    &0.423 &0.438 &0.456 &0.466 &0.483\\
\midrule
\multirow{2}{*}{{Informer}}     &$MSE(\downarrow)$ 
&0.577 &0.685 &0.931 &1.128 &1.215    &0.720 &1.457 &3.489 &2.723 &3.467
&0.323 &0.494 &0.678 &1.056 &1.192    &0.312 &0.392 &0.515 &0.759 &0.969   \\
&$MAE(\downarrow)$ 
&0.549 &0.625 &0.752 &0.873 &0.896            &0.665 &1.001 &1.515 &1.340 &1.473
&0.369 &0.503 &0.614 &0.786 &0.926            &0.387 &0.431 &0.509 &0.625 &0.788 \\
\midrule                          
\multirow{2}{*}{{TS2Vec}} &$MSE(\downarrow)$
&0.599 &0.629 &0.755 &0.907 &1.048       &0.398 &0.580 &1.901 &2.304 &2.650
&0.443 &0.582 &0.622 &0.709 &0.786       &0.287 &0.307 &0.332 &0.349 &0.375\\
& $MAE(\downarrow)$   
&0.534 &0.555 &0.636 &0.717 &0.790      &0.461 &0.573 &1.065 &1.215 &1.373
&0.436 &0.515 &0.549 &0.609 &0.655      &0.374 &0.388 &0.407 &0.420 &0.438\\  
\midrule  
\midrule
\multirow{2}{*}{\textbf{DE-TSMCL}}     &$MSE(\downarrow)$ 
&\textbf{0.569} &\textbf{0.620} &\textbf{0.744} &\textbf{0.899} &\textbf{1.002}    
&\textbf{0.376} &\textbf{0.564} &\textbf{1.818} &\textbf{2.120} &\textbf{2.376}     
&\textbf{0.391} &\textbf{0.549} &\textbf{0.601} &\textbf{0.660}  & \textbf{0.741}    
&\textbf{0.250} &\textbf{0.292} &\textbf{0.303} &\textbf{0.337} &\textbf{0.333}  \\
& $MAE(\downarrow)$ 
&\textbf{0.524} &\textbf{0.548} &\textbf{0.623} &\textbf{0.706} &\textbf{0.776}
 &\textbf{0.454} &\textbf{0.565} &\textbf{1.058} &\textbf{1.213}  &\textbf{ 1.270  }
 &\textbf{0.401} &\textbf{0.468} &\textbf{0.503} &\textbf{0.571}  &\textbf{0.611}  
 &\textbf{0.269} &\textbf{0.299} &\textbf{0.372} &\textbf{0.408} &\textbf{0.424}  \\
\bottomrule
\end{tabular}
}\end{table*}

\subsection{\textbf{Main Results(RQ1)}} 
Table \ref{comparison_uni} and Table \ref{comparison_mu} report the comparison results of different approaches for univariate time series and multivariate time series forecasting, respectively. We will analyze the experimental results in the rest of the section.

\begin{table*}[]
\centering
\caption{Comparison results of different approach for univariate and multivariate forecasting on ETTm2 dataset.}
\label{m2}
\scalebox{1.0}{ 
\begin{tabular}{lcccccccccc}
\toprule
~  & \multirow{2}{*}{P} & \multicolumn{2}{c}{DE-TSMCL} & \multicolumn{2}{c}{Informer} & \multicolumn{2}{c}{LogTrans} & \multicolumn{2}{c}{Reformer} \\
 ~   &\multirow{1}{*}{}  & MSE & MAE & MSE & MAE  & MSE & MAE & MSE & MAE \\
\midrule
\multirow{4}{*}{Univariate}
            & 96   & \textbf{0.086} & \textbf{0.199}  &0.088 & 0.225 &0.075 & 0.208 &0.131 & 0.288 \\
            & 192  & \textbf{0.118} & \textbf{0.251} &0.132 & 0.283 &0.129 & 0.275 &0.186 & 0.354 \\
           & 336  & \textbf{0.152} & \textbf{0.301} &0.180 & 0.336 &0.154 & 0.302 &0.220 & 0.381 \\
           & 720  & 0.200 & \textbf{0.319} &0.300 & 0.435 &\textbf{0.160} & 0.321 &0.267 & 0.430 \\
\midrule
\multirow{4}{*}{Multivariate}
             & 96   & \textbf{0.304} & \textbf{0.356} &0.355 & 0.462 &0.768 & 0.642 &0.658 & 0.619 \\
             & 192  & \textbf{0.315} & \textbf{0.369}  &0.595 & 0.586 &0.989 &      0.757 &1.078 & 0.827\\
             & 336  & \textbf{0.337} & \textbf{0.381} &1.270 & 0.871 &1.334 & 0.872 &1.549 & 0.972\\
             & 720  & \textbf{0.398} & \textbf{0.415} &3.001 & 1.267 &3.048 & 1.328 &2.631 & 1.242 \\
\bottomrule
\end{tabular}
}\label{time}
\end{table*}

\subsubsection{\textbf{Univariate Time Series Forecasting}}
The results for univariate time series forecasting of ten different methods for four datasets are tabulated in Table\ref{comparison_uni}. On all the different datasets, {DE-TSMCL} consistently obtains the best performance under two evaluation metrics. In particular, when compared with the up-to-date contrastive learning based method TS2Vec, {DE-TSMCL} achieves obvious performance improvement. For example, {DE-TSMCL} achieves 24.2\% improvement in MSE and 14.7\% in MAE on the ETTm1 dataset, which contains a considerable volume of data. On the ETTh1 dataset, {DE-TSMCL} also gets similar gains(14.2\% improvement in MSE and 10.6\% in MAE). Such significant performance gains are primarily attributed to momentum update and distillation enhanced  learning framework. A larger dataset can benefit the training process of both the teacher and student networks in a knowledge distillation setup.
When the dataset constains a large volume of data, it allows the teacher network to undergo more updates, refining its knowledge and providing a more accurate target distribution for the student. Thus, the joint training of the teacher and student networks can lead to improved representation and prediction results by leveraging the teacher's expertise and the diversity of the data.

\subsubsection{\textbf{Multivariate Time Series Forecasting}}
The results for multivariate time series forecasting of nine methods for four datasets are reported in Table \ref{comparison_mu}. It is easy to observe that the proposed {DE-TSMCL} model consistently outperforms other state-of-the-art approaches on all datasets. These results indicate that the pre-trained encoder effectively captures the features of time series to improve prediction  performance. Similar to univariate time series forecasting, {DE-TSMCL} achieves remarkable improvements on ETTm1 and electricity datasets. Specifically, we achieve a maximum improvement of 12.2\% on the MSE criterion and 27.3\% on the MAE criterion. 

To provide additional experimental results in the energy domain, we conduct experiments for both univariate and multivariate forecasting using the ETTm2 dataset. Table \ref{m2} presents the comparison results of different approaches.
We observe that our model consistently demonstrates improved performance as the prediction length increases. Specifically, the accuracy increased from 3.3\textperthousand to 6.2\textperthousand  when the prediction length extends from 336 to 720 time steps for univariate forecasting. This indicates that our model becomes more accurate with longer prediction horizons.
Furthermore, we note that the improvement in multivariate forecasting is greater compared to univariate time series forecasting. This observation aligns with the findings from the previous section, which indicates that the momentum updating technique benefits more from having more training data.

\begin{table}[]
\caption{Ablation results with prediction lengths $P\in \{24, 68, 168, 336, 720\}$ for ETTh1.  Results are averaged from all prediction lengths.}
    \label{abh1}
    \centering
    \begin{tabular}{cccccccc}
    \toprule 
        ~ &\multirow{2}{*}{M}  &\multirow{2}{*}{C} &\multirow{2}{*}{S} & \multicolumn{2}{c}{Univariate} &  \multicolumn{2}{c}{Multivariate}  
        \\ \cline{5-8}
        
        ~ &\multirow{2}{*}{} & \multirow{2}{*}{} & \multirow{2}{*}{} & MSE & MAE & MSE & MAE \\ \hline
        Basic & ~ & ~ & ~ & 0.106 & 0.245 & 0.790 & 0.653 \\ \hline
        Basic+C & ~ &\ding{51} &~ & 0.105 & 0.246  & 0.792 & 0.645 \\ \hline
        Basic+S & ~ & ~ & \ding{51} & 0.103 & 0.242 & 0.779 & 0.645 \\ \hline
        Basic+C+S & ~ & \ding{51} & \ding{51} & 0.103 & 0.239 & 0.781 & 0.641 \\ \hline
        DE+M & \ding{51} &~ & ~ & 0.103 & 0.240 & 0.787 & 0.652 \\ \hline
        DE+M+C & \ding{51} & \ding{51} & ~ & 0.102 & 0.238 & 0.785 & 0.647 \\ \hline
        DE+M+S & \ding{51} &~ & \ding{51} & 0.102 &\textbf{0.236} & 0.770 & 0.639 \\ \hline
        DE-TSMCL & \ding{51} & \ding{51} & \ding{51} &\textbf{0.101} &\textbf{0.236} &\textbf{0.767} &\textbf{0.635} \\
        \bottomrule 
    \end{tabular}
\end{table}

\begin{table}[]
\caption{Ablation results with prediction lengths $P\in \{24, 68, 168, 336, 720\}$ for ETTh2.  Results are averaged from all prediction lengths.}
    \label{abh2}
    \centering
    \begin{tabular}{cccccccc}
    \toprule 
        ~ &\multirow{2}{*}{M}  &\multirow{2}{*}{C} &\multirow{2}{*}{S} & \multicolumn{2}{c}{Univariate} &  \multicolumn{2}{c}{Multivariate}  
        \\ \cline{5-8}
        
        ~ &\multirow{2}{*}{} & \multirow{2}{*}{} & \multirow{2}{*}{} & MSE & MAE & MSE & MAE \\ \hline
        Basic & ~ & ~ & ~ & 0.166 & 0.319 &1.571 & 0.941 \\ \hline
        Basic+C & ~ &\ding{51}&~ & 0.166 & 0.319  &1.506 & 0.927 \\ \hline
        Basic+S & ~ & ~ & \ding{51} & 0.164 & 0.317 & \textbf{1.451} & {0.916} \\ \hline
        Basic+C+S &~ & \ding{51} & \ding{51} & 0.164 & 0.316 & 1.454 & 0.917 \\ \hline
        DE+M & \ding{51} & ~ & ~ & 0.165 & 0.316 & 1.457 & 0.919 \\ \hline
        DE+M+C & \ding{51} & \ding{51} & ~ & 0.165 & \textbf{0.315} & 1.460 & 0.915 \\ \hline
        DE+M+S &\ding{51} & ~ & \ding{51} & 0.161 & \textbf{0.315} & 1.452 & 0.913 \\ \hline
        DE-TSMCL & \ding{51} & \ding{51} & \ding{51} &\textbf{0.159} &\textbf{0.315} &\textbf{1.451} &\textbf{0.912} \\
        \bottomrule 
    \end{tabular}
\end{table}

\subsection{Ablation Study(RQ2)} 
In this section, we conduct extensive ablation studies to validate the effectiveness of key components in DE-TSMCL, including distillation framework (DE), momentum update (M), center layer (C) and supervised task (S). we design seven variants of DE-TSMCL. It is worth noting that we have added each component and their combination on the basic model to examine the importance of each component. 
\begin{itemize}
    \item \textbf{Basic}: We use two independent encoders with the same architecture to generate separate representations of time series data, and the loss function is the ordinary InfoNCE. 
    \item \textbf{Basic+C}: We add a center layer based on the basic model.
    \item \textbf{Basic+S}: We add a supervised-task based on the basic model.
    \item \textbf{Basic+C+S}: We add both a center layer and a supervised task-based on the basic model.
    \item \textbf{DE+M}: We incorporate the knowledge distillation framework with momentum update based on the basic model.
    \item \textbf{DE+M+C}: We add  a center layer based on the distillation enhanced framework.
    \item \textbf{DE+M+S}: We add  a supervise task based on the distillation enhanced framework.
\end{itemize} 

Table \ref{abh1} and  Table \ref{abh2} report the experimental results of all variants for ETTh1 and ETTh2 datasets, respectively. 
We start from the basic forecasting framework and introduce each component or their combination in turn. As shown in the table, adopting distillation enhanced
framework(DE+M) has provided 2.9\% (MSE) and +2.1\% (MAE) improvement for univariate time series forecasting, as well as 0.4\% (MSE) and 0.2\% (MAE) improvement for multivarite time series forecasting on ETTh1.
For ETTh2 dataset, the improvement is up to 7.3\% (MSE) and 2.4\% (MAE) for multivarite time series forecasting. Moreover, when combing with other components, the improvement can increase to 7.1\% (DE+M+C), 7.5\% (DE+M+S) and 7.7\% (DE-TSMCL). Theses results demonstrate that our proposed distillation enhanced framework with momentum update is able to obviously improve the representation capability, moreover help achieve better forecasting performance than the basic time series prediction method.

\begin{figure*}
\centering
 \begin{minipage}{0.237\linewidth}
  \centerline{\includegraphics[width=\textwidth]{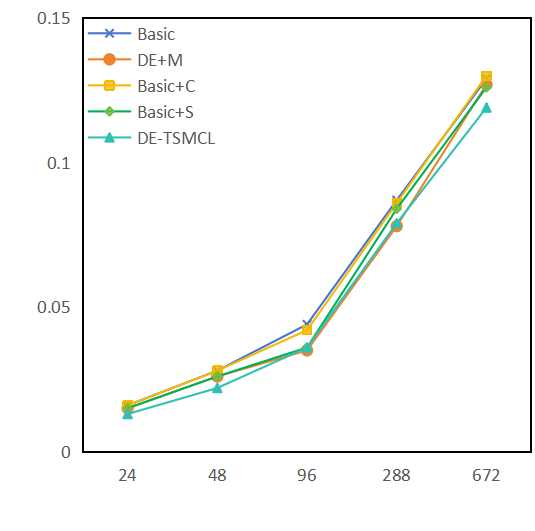}}
  \centerline{(a) MSE on ETTm1}
 \end{minipage}
 \begin{minipage}{0.237\linewidth}
  \centerline{\includegraphics[width=\textwidth]{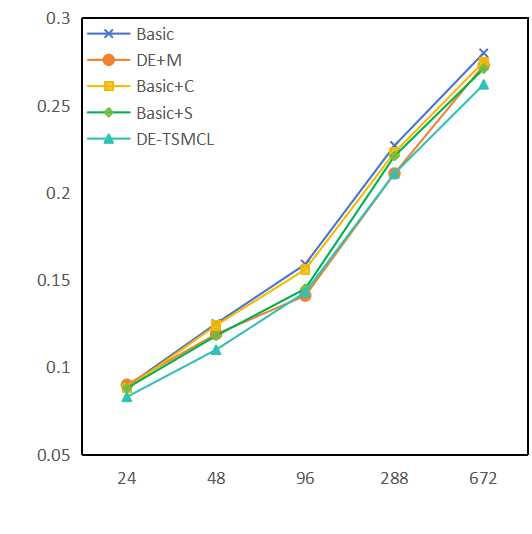}}
  \centerline{(b) MAE on ETTm1}
 \end{minipage} 
 \begin{minipage}{0.237\linewidth}
   \centerline{\includegraphics[width=\textwidth]{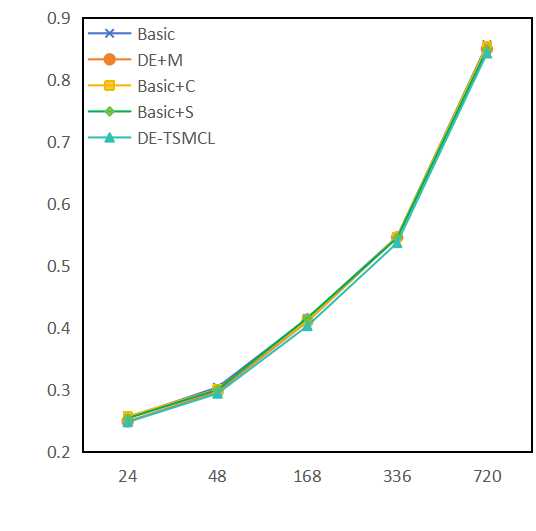}}
  \centerline{(c) MSE on Electricity}
 \end{minipage}
 \begin{minipage}{0.237\linewidth}
  \centerline{\includegraphics[width=\textwidth]{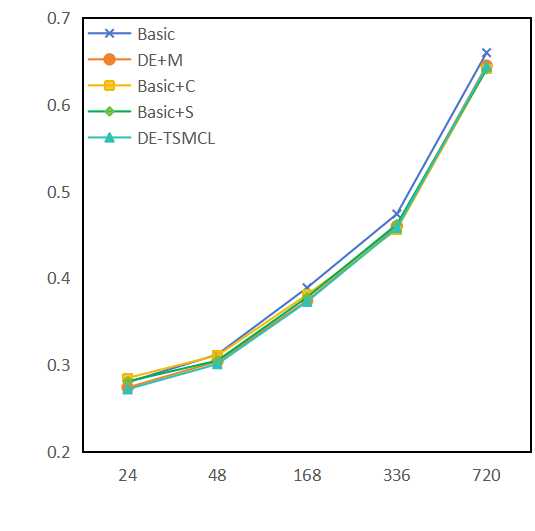}}
  \centerline{(d) MAE on Electricity}
 \end{minipage}
 
\caption{The effect of each component of DE-TSMCL for univariate time series forecasting.}
\label{single-graph}
\label{m1eleuni}
\end{figure*}

\begin{figure*}\centering
 \begin{minipage}{0.237\linewidth}
    \centerline{\includegraphics[width=\textwidth]{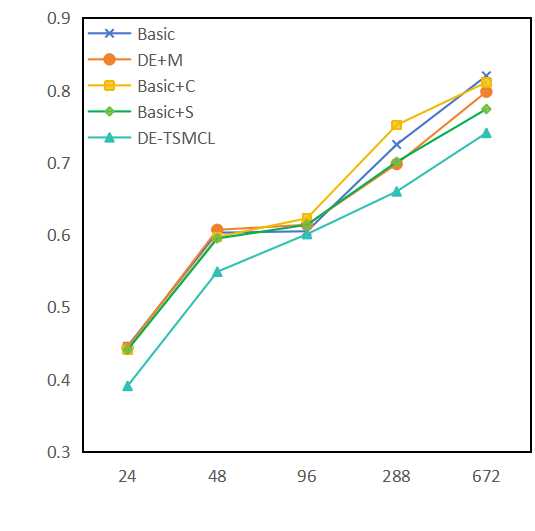}}
  \centerline{(a) MSE on ETTm1}
 \end{minipage}
 \begin{minipage}{0.237\linewidth}
  \centerline{\includegraphics[width=\textwidth]{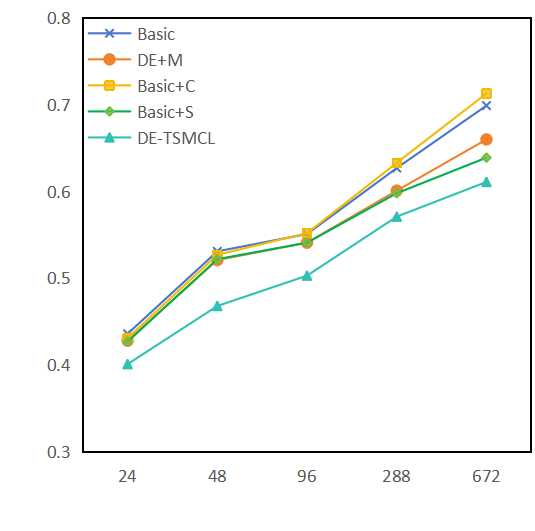}}
  \centerline{(b) MAE on ETTm1}
 \end{minipage}
 \begin{minipage}{0.237\linewidth}
  \centerline{\includegraphics[width=\textwidth]{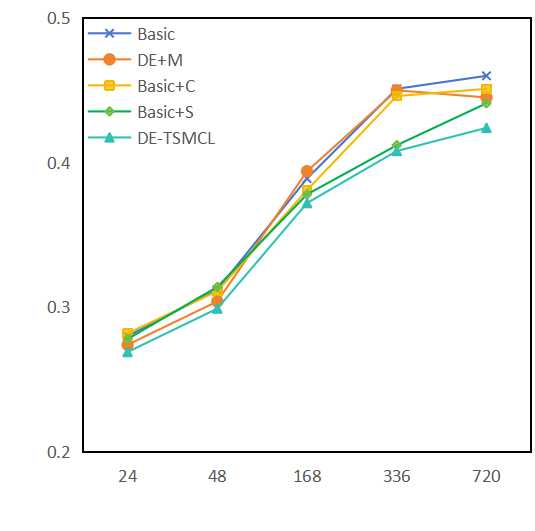}}
  \centerline{(c) MSE on Electricity}
 \end{minipage}
 \begin{minipage}{0.237\linewidth}
  \centerline{\includegraphics[width=\textwidth]{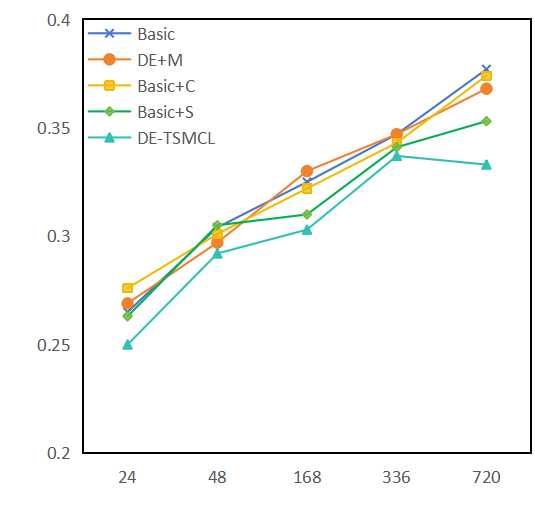}}
  \centerline{(d) MAE on Electricity}
 \end{minipage}
\caption{The effect of each component of DE-TSMCL for multivariate time series forecasting.}
\label{multi-graph}
\label{m1elemu}
\end{figure*}

In order to visualize the impact of each component on all prediction , we plot Fig. \ref{m1eleuni} and Fig. \ref{m1elemu} to visually present the forecasting results of four basic variants for {ETTm1} and {electricity} datasets. We can observe that the performance improves by incorporating distillation framework with momentum update, supervised task and center layer, which demonstrates that the designed framework is effective to enhance the representation ability of the model to improve the prediction performance. First, it is obvious that the performance of Basic/Basic+C/Basic+S is worse than DE-TSMCL, indicating the advantage of utilizing the distillation framework with momentum update for learning representations of time series data. Moreover, the advantage exhibits variations across different datasets. The main reason is that  ETTm1 contains greater volume of data, which achieves more benefit from the training process of both the teacher and student networks in a knowledge distillation setup.
Second, the performance of DE+M is also inferior to DE-TSMCL, which invalidates the function of supervised task. In addition, it is observed that the inclusion of only center layer (Basic+C) or supervised task(Basic+S) also slightly improves the performance of our model. Therefore, the superior performance of DE-TSMCL demonstrates the effectiveness of our design in introducing distillation framework with momentum update and supervised task for time series forecasting.

\subsection{Hyperparameter Analysis(RQ3)}
In this section, we investigate the sensitivity of DE-TSMCL about several key hyperparameters: the weight for self-supervised loss $\lambda$ and the momentum coefficient $m$. 

\begin{figure*}
\centering
     \begin{minipage}{0.237\linewidth}
  \centerline{\includegraphics[width=\textwidth]{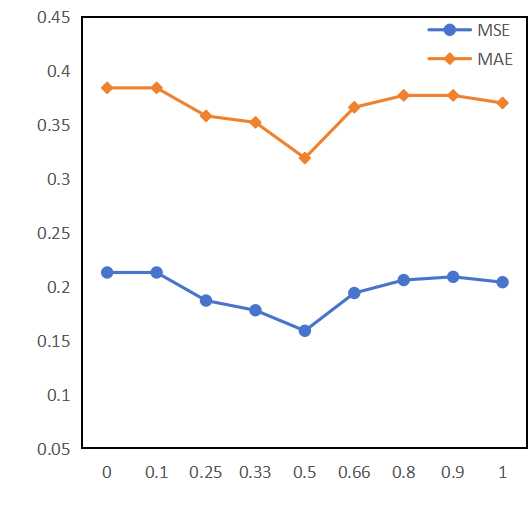}}
  \centerline{(a) ETTh1}
 \end{minipage}
  \begin{minipage}{0.237\linewidth}
  \centerline{\includegraphics[width=\textwidth]{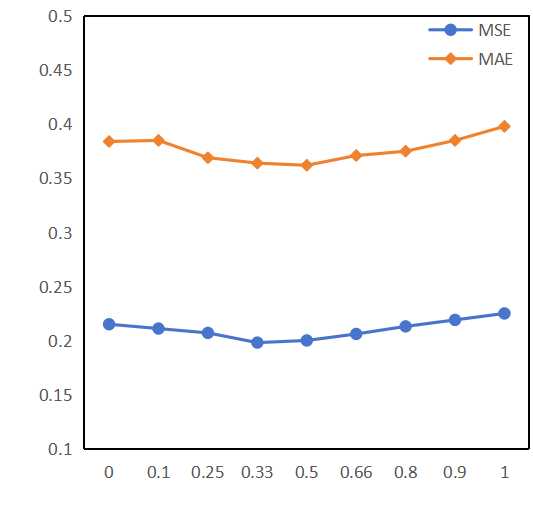}}
  \centerline{(b) ETTh2}
 \end{minipage}
   \begin{minipage}{0.237\linewidth}
  \centerline{\includegraphics[width=\textwidth]{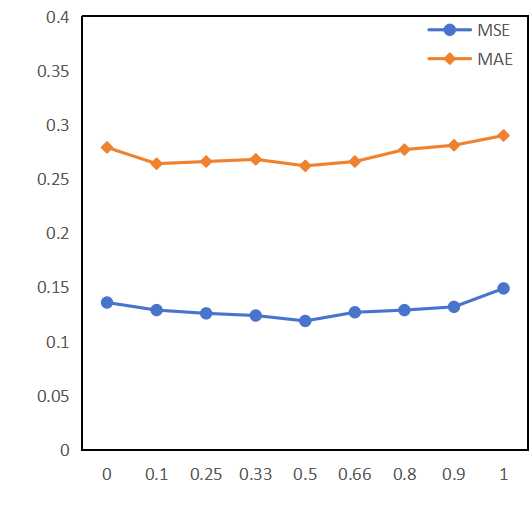}}
  \centerline{(c) ETTm1}
 \end{minipage}
   \begin{minipage}{0.237\linewidth}
  \centerline{\includegraphics[width=\textwidth]{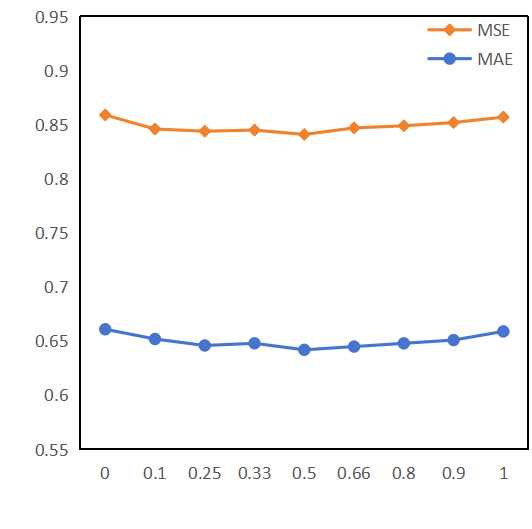}}
  \centerline{(d) Electricity}
 \end{minipage}
 \caption{The impact of $\lambda$ on four different datasets for univariate time series forecasting.}
    \label{hyperl}
\end{figure*}

\begin{figure*}
\centering
     \begin{minipage}{0.237\linewidth}
  \centerline{\includegraphics[width=\textwidth]{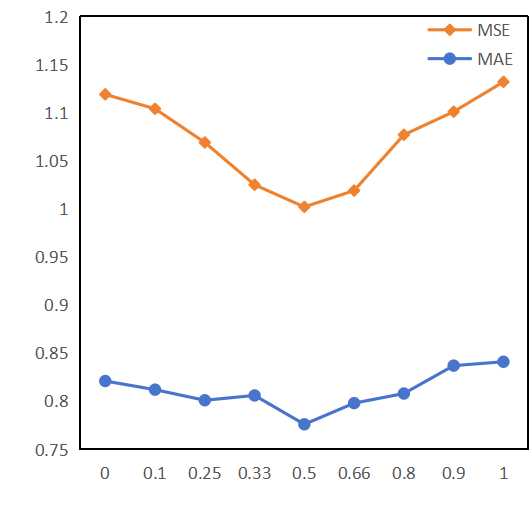}}
  \centerline{(a) ETTh1}
 \end{minipage}
  \begin{minipage}{0.237\linewidth}
  \centerline{\includegraphics[width=\textwidth]{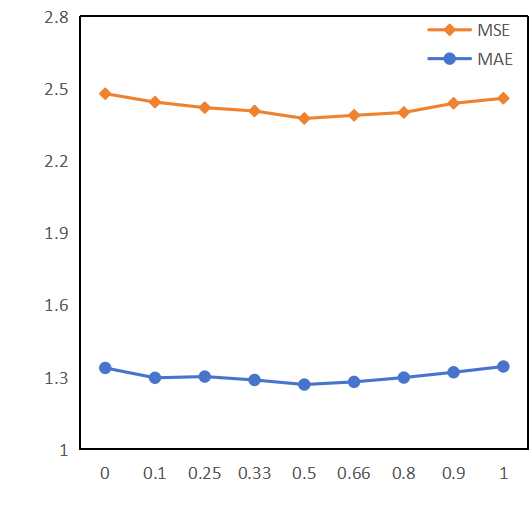}}
  \centerline{(b) ETTh2}
 \end{minipage}
   \begin{minipage}{0.237\linewidth}
  \centerline{\includegraphics[width=\textwidth]{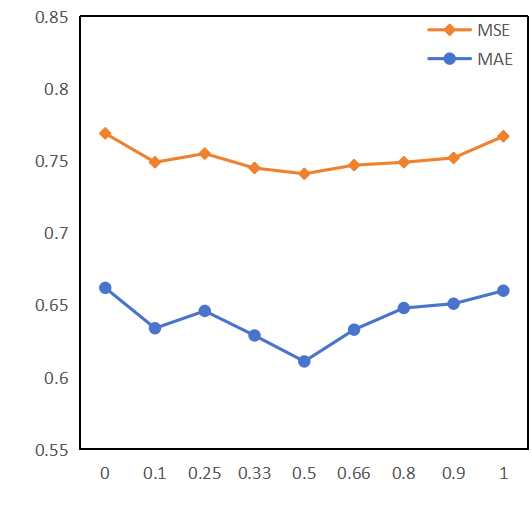}}
  \centerline{(c) ETTm1}
 \end{minipage}
   \begin{minipage}{0.237\linewidth}
  \centerline{\includegraphics[width=\textwidth]{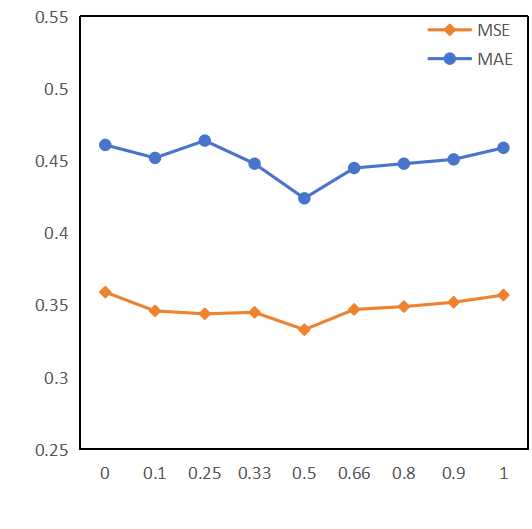}}
  \centerline{(d) Electricity }
 \end{minipage}
\caption{The impact of $\lambda$ on four different datasets for multivariate time series forecasting.}
\label{hyperlm}
\end{figure*}

\subsubsection{The Impact of $\lambda$}
As our model jointly optimizes the supervised and self-supervised tasks with hyperparameter $\lambda$ in Eq.(\ref{joint}), we first explore the effect of $\lambda$ on the model performance. 
The value of $\lambda$ determines the importance of the self-supervised task relative to the supervised task, we tune it in $\{0,0.1,0.25,0.33,0.5,0.66,0.8,0.9,1.0\}$ to find the best balance for optimal results. Fig.\ref{hyperl} and \ref{hyperlm} illustrate the results on four datasets for univariate and multivariate forecasting, respectively. It is observed that as $\lambda$ increases from $0$ to $0.5$, the prediction accuracy increases significantly and peaks at $\lambda = 0.5$. This suggests that the ratio of self-supervised and supervised tasks is optimal at this point. However, when $\lambda >0.5$, the performance tends to decrease. The low proportion of self-supervised learning has an impact on the model's performance, which is the reason for the observed results. These findings confirm the advantage of jointly optimizing both supervised and self-supervised tasks.

\subsubsection{The Impact of $m$}
\begin{figure*}
\centering
     \begin{minipage}{0.241\linewidth}
  \centerline{\includegraphics[width=\textwidth]{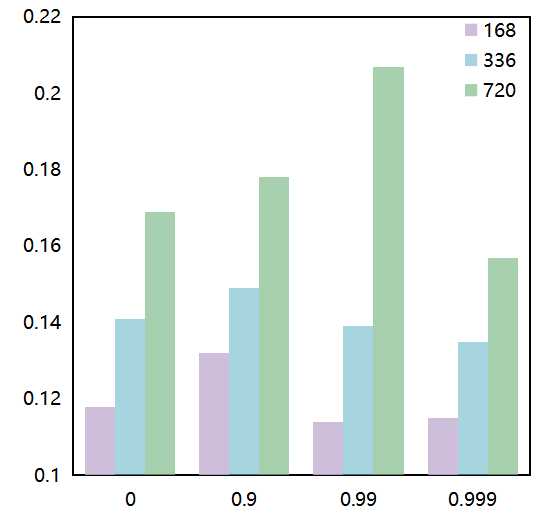}}
  \centerline{(a) MSE on ETTh1}
 \end{minipage}
  \begin{minipage}{0.241\linewidth}
  \centerline{\includegraphics[width=\textwidth]{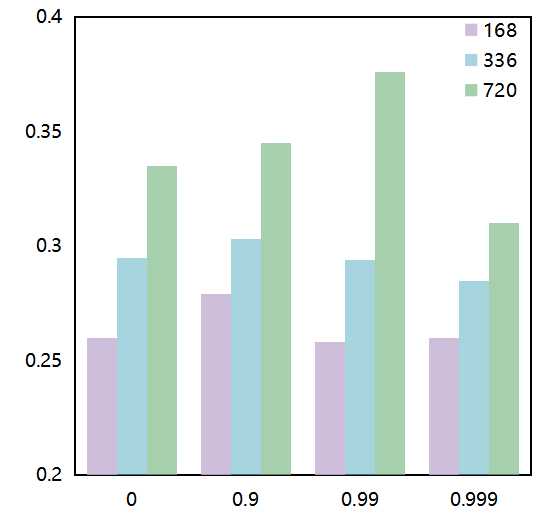}}
  \centerline{(b) MAE on ETTh1 }
 \end{minipage}
   \begin{minipage}{0.241\linewidth}
  \centerline{\includegraphics[width=\textwidth]{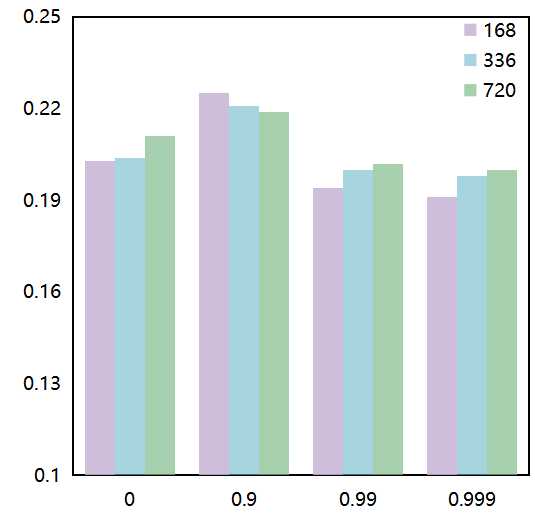}}
  \centerline{(c) MSE on ETTh2 }
 \end{minipage}
   \begin{minipage}{0.241\linewidth}
  \centerline{\includegraphics[width=\textwidth]{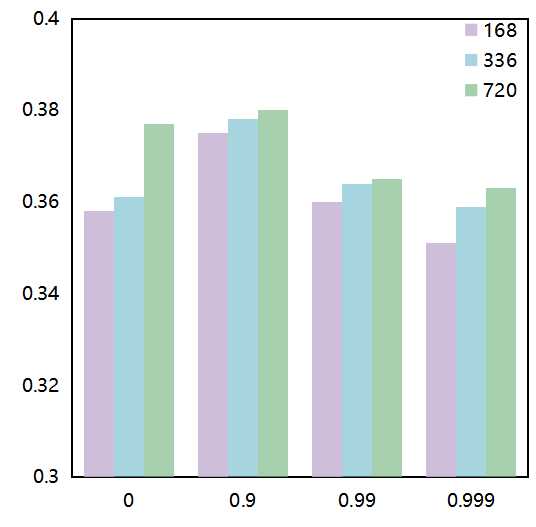}}
  \centerline{(d) MAE on ETTh2}
 \end{minipage}
    \caption{The impact of $m$ on ETTh1 and ETTh2 datasets for univariate forecasting ($l$=168, 336, 720).}
    \label{hyperm}
\end{figure*}

\begin{figure*}
\centering
     \begin{minipage}{0.241\linewidth}
 \centerline{\includegraphics[width=\textwidth]{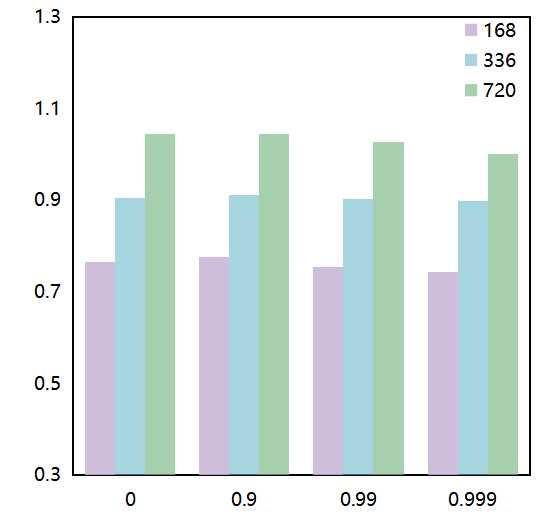}}
  \centerline{(a) MSE on ETTh1}
 \end{minipage}
  \begin{minipage}{0.241\linewidth}
\centerline{\includegraphics[width=\textwidth]{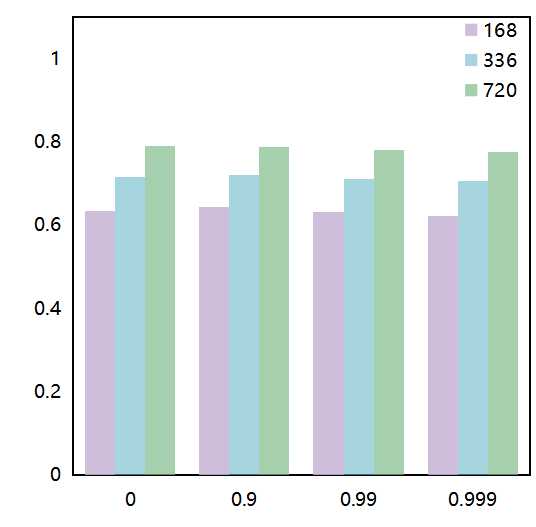}}
  \centerline{(b) MAE on ETTh1 }
 \end{minipage}
   \begin{minipage}{0.241\linewidth}
  \centerline{\includegraphics[width=\textwidth]{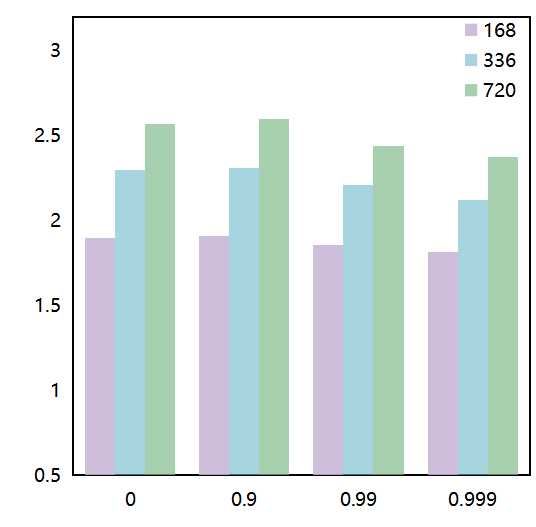}}
  \centerline{(c) MSE on ETTh2 }
 \end{minipage}
   \begin{minipage}{0.241\linewidth}
  \centerline{\includegraphics[width=\textwidth]{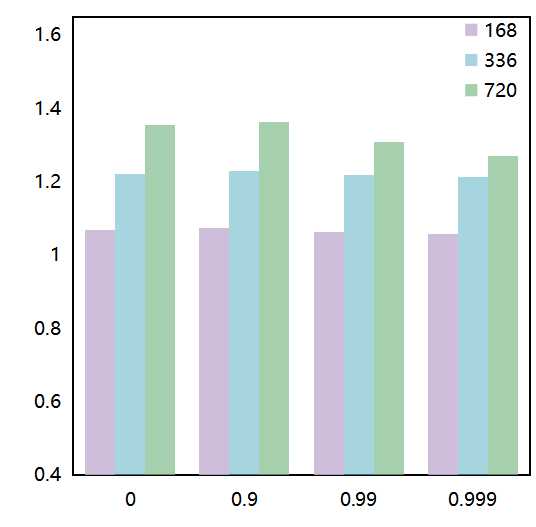}}
  \centerline{(d) MAE on ETTh2}
 \end{minipage}
\caption{The impact of $m$ on ETTh1 and ETTh2 datasets for multivariate time series forecasting ($l$=168, 336, 720).}
\label{hypermm}
\end{figure*}

In this section, we evaluate the influence of momentum coefficient on the forecasting performance.
Figure \ref{hyperm} and \ref{hypermm} depict the impact of different values of $m \in {0,0.9,0.99,0.999}$ on the varying sizes of prediction length for the ETTh1 and ETTh2 datasets. It can be observed that as $m$ approaches 1.0, the experimental results consistently improve. The best results are achieved when $m=0.999$, which aligns with the findings in the case of CV\cite{he2020momentum}. The reason is that our model incorporates momentum contrast with a slowly updating encoder parameter, which promotes a more stable learning process. This stability is especially crucial in time series forecasting tasks, where maintaining consistency and continuity in the learned representations is essential for accurate predictions.

\subsection{Visualization Analysis} 
\begin{figure*}
\centering
 \begin{minipage}{0.44\linewidth}
     \label{visual-my720}
  \centerline{\includegraphics[width=\textwidth]{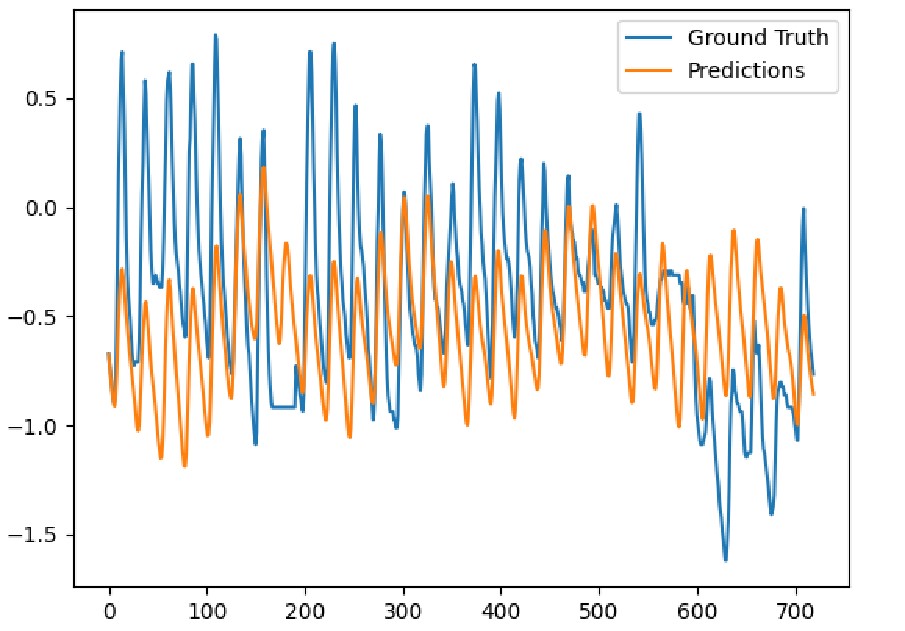}} 
  \centerline{(a) Visualizing the DE-TSMCL prediction (P=720).}
 \end{minipage}
  \begin{minipage}{0.44\linewidth}
     \label{visual-ts 720}
  \centerline{\includegraphics[width=\textwidth]{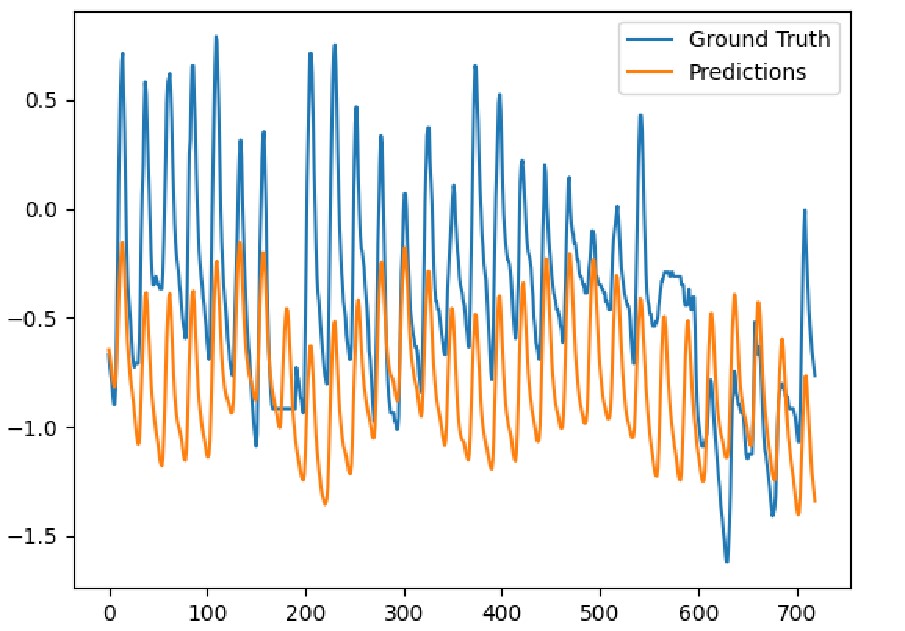}}
  \centerline{(b) Visualizing the TS2vec prediction (P=720).}
 \end{minipage}
  \begin{minipage}{0.44\linewidth}
     \label{visual-my110}
  \centerline{\includegraphics[width=\textwidth]{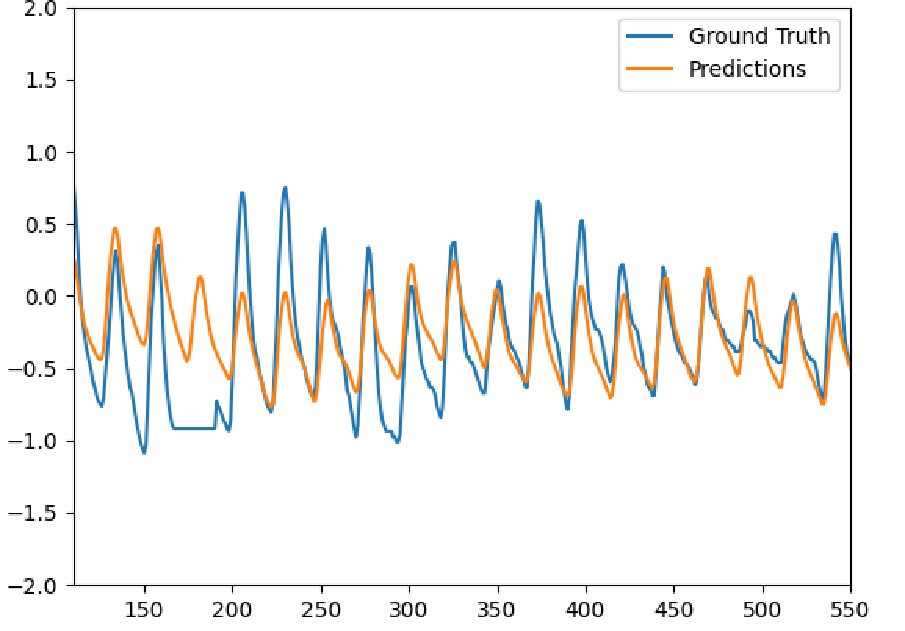}}
  \centerline{(c) Visualizing the DE-TSMCL part prediction.}
 \end{minipage}
  \begin{minipage}{0.44\linewidth}
     \label{visual-ts 110}
  \centerline{\includegraphics[width=\textwidth]{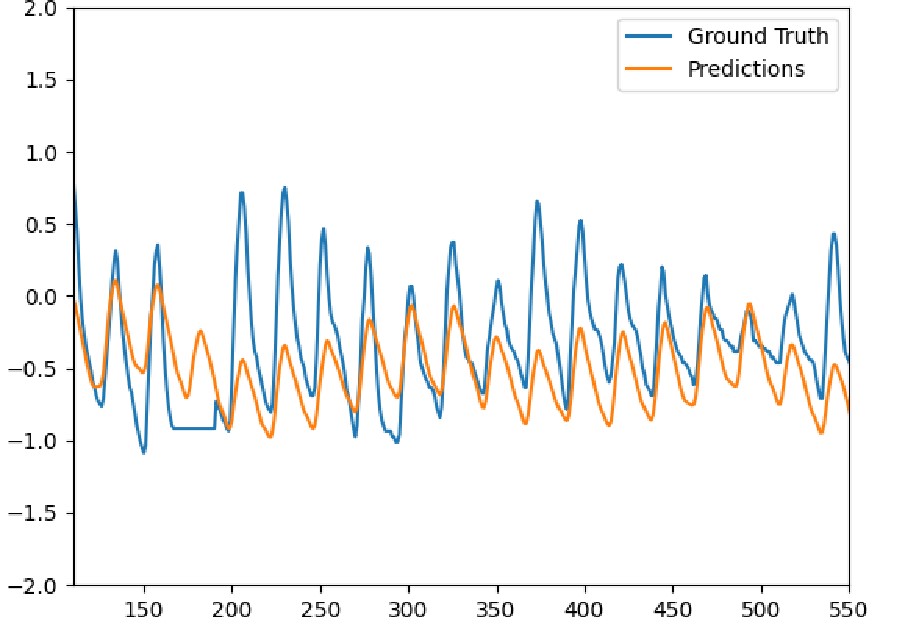}}
  \centerline{(d) Visualizing the TS2vec part prediction.}
 \end{minipage}
\caption{Visualisation results of DE-TSMCL and TS2vec for long-term predicting on ETTh2.}
\label{Visual}
\end{figure*}

\begin{figure*}
 \vspace{+0.2cm} 
\centering
 \begin{minipage}{0.44\linewidth}
\centerline{\includegraphics[width=\textwidth]{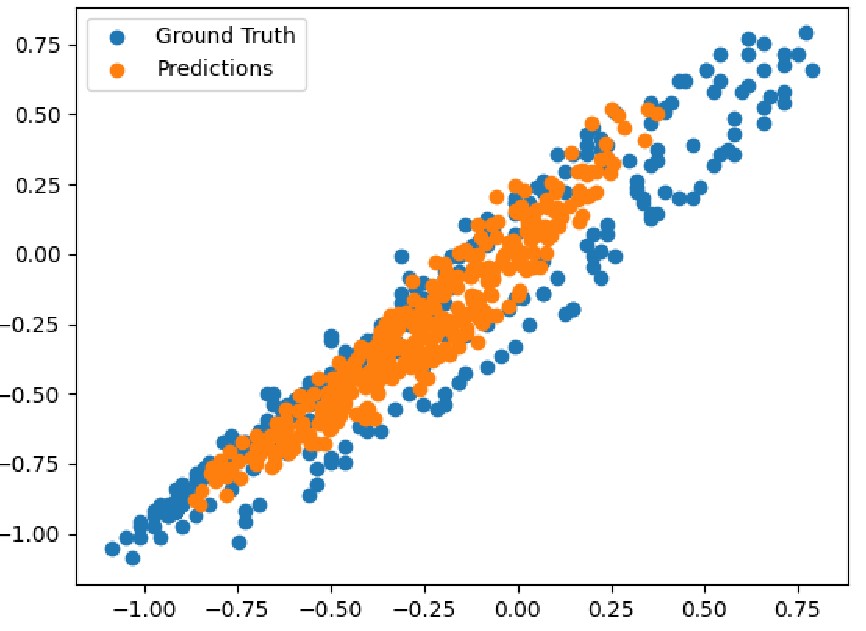}}
  \centerline{(a) Complete scatter plot.}
 \end{minipage}
 \begin{minipage}{0.46\linewidth}
\centerline{\includegraphics[width=\textwidth]{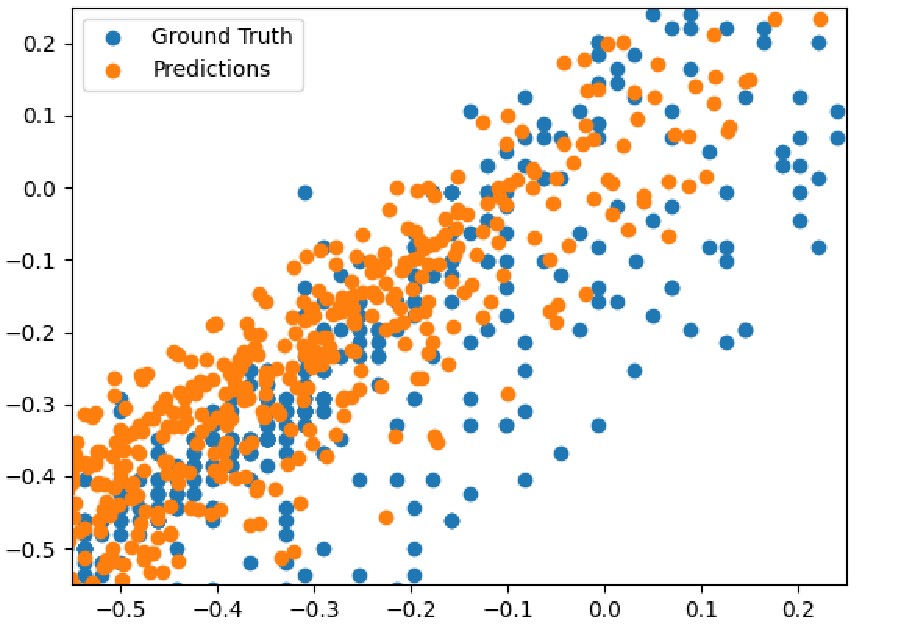}}
  \centerline{(b) Partial scatter plot.}
 \end{minipage}
\caption{Visualisation results of DE-TSMCL for long-term prediction on ETTh2.}
\label{sdt}
\end{figure*}

To further evaluate the forecasting quality of the proposed DE-TSMCL, we present an additional visualization analysis. 
We select time serie in ETTh2 to visualize the differences between ground truth and values predicted by DE-TSMCL as shown in Fig.\ref{Visual}. Additionally, we plot the prediction results of TS2Vec, which performs the best among the baseline methods, for comparison.
In order to demonstrate the robustness of our model to shifts in time series distribution and anomalous changes, we have zoomed in on specific sections of the visualizations. Figure 9(c) and Figure 9(d) provide a closer look at these selected sections. To further analyze and compare the differences, we create scatter plots of Figure 9(a) and Figure 9(c), which are shown in Figure \ref{sdt}.

Our observations from the visual analysis are as follows:(1)
The ground truth curve exhibits irregular and wild fluctuations at certain points. However, DE-TSMCL demonstrates the ability to fit the curve accurately and provides stable and precise forecasting performance. This indicates that our model can effectively capture the complex dynamics and fluctuations present in the time series data. (2)
When comparing DE-TSMCL with the state-of-the-art model, TS2Vec, we observe that DE-TSMCL showcases a more precise modeling ability. This suggests that our proposed model, which integrates self-supervised and supervised tasks with momentum contrastive learning, is effective and efficient in terms of time series representation and prediction.



\begin{table*}[]
\centering
\caption{The effect of data augmentation for univariate time series forecasting.}
\label{ablation_spre}
\scalebox{0.55}
{
\begin{tabular}{ll|ccccc|ccccc|ccccc|ccccc}
\toprule
\multicolumn{2}{l|}{Datasets}        & \multicolumn{5}{c|}{ETTh1} & \multicolumn{5}{c|}{ETTh2} & \multicolumn{5}{c|}{ETTm1} & \multicolumn{5}{c}{Electricity} \\
\midrule
Methods & Metrics &24 &48 &168 &336 &720  &24 &48 &168 &336 &720   &24 &48 &96 &288 &672  &24     &48     &168     &336  &720  \\
\midrule 
\multirow{2}{*}{\textbf{DE-TSMCL-pre}}     &$MSE(\downarrow)$ 
&0.039 &0.066 &0.132 &0.140  &\underline{0.162}
     &\textbf{0.086} &\underline{0.121} &0.202 &0.210 &0.209
 &\underline{0.015} &0.028 &0.040 &0.081 &\underline{0.123}
      &\underline{0.249} &{0.296} &\underline{0.408} &\underline{0.545}  &\underline{0.847}\\
& $MAE(\downarrow)$ 
&\underline{0.148} &0.184 &0.280 &0.298 &\underline{0.322}
&\textbf{0.224} &\textbf{0.267} &0.362 &0.367 &0.371    
&0.091 &0.125 &0.149 &\textbf{0.209} &\underline{0.266}
&0.275 &\underline{0.302} &\underline{0.373} &0.463 &\textbf{0.641}\\
\midrule
\multirow{2}{*}{\textbf{DE-TSMCL-after}}     &$MSE(\downarrow)$ 
&\underline{0.039} &\underline{0.060} &\underline{0.113} &\underline{0.137}  & 0.164   
&0.088 &{0.122} &{0.201} &{0.206} &{0.208}
&0.015 &\underline{0.026} &\underline{0.035} &\textbf{0.078} &0.127
&\underline{0.249} &{0.297} &0.410&{0.546} &0.850 \\
                          & $MAE(\downarrow)$ 
 &{0.149} &\underline{0.181} &\textbf{0.252} &\underline{0.289} &0.330
 &\underline{0.227} &{0.272} &\underline{0.350} &{0.365} &\underline{0.368}        
 &\underline{0.090} &\underline{0.119} &\textbf{0.141} &{0.211} &0.273     
 &{0.274} &{0.304} &0.374 &\underline{0.460} &0.645  \\
\midrule

\multirow{2}{*}{\textbf{DE-TSMCL-reverse}}     
&$MSE(\downarrow)$ 
&{0.042} &{0.065} &{0.142} &{0.148} &{0.166}    
&{0.090} &{0.129} &\underline{0.194} &\underline{0.199} &\underline{0.202}    
&{0.016} &{0.027} &{0.040} &{0.083}  &{0.131}    
&{0.248} &{0.293} &{0.403} &{0.553} &{0.897}  \\
& $MAE(\downarrow)$ 
&{0.152} &{0.182} &{0.270} &{0.295} &{0.337}
 &{0.231} &\underline{0.269} &{0.342} &{0.357}  &\underline{0.366}
 &{0.095} &{0.123} &{0.150} &{0.217}  &{0.278}  
 &\underline{0.273} &\underline{0.302} &{0.371} &{0.461} &\underline{0.644}  \\
\midrule

\multirow{2}{*}{\textbf{DE-TSMCL}}     &$MSE(\downarrow)$ 
&{0.038} &{0.059} &{0.115} &{0.135} &{0.157}  
&\underline{0.088} &\textbf{0.120} &\textbf{0.191} &\textbf{0.198} &\textbf{0.200}        
&\textbf{0.013} &\textbf{0.022} &\textbf{0.036} &\underline{0.079}  &\textbf{0.119}    
&\textbf{0.248} &\underline{0.294} &\textbf{0.403} &\textbf{0.537} &\textbf{0.843}  \\
& $MAE(\downarrow)$ 
&\textbf{0.146} &\textbf{0.181} &\underline{0.260} &\textbf{0.285} &\textbf{0.310}
&{0.229} &{0.272} & {0.351} &\underline{0.359}  
&\textbf{0.363}  
&\textbf{0.083} &\textbf{0.110} &\underline{0.143} &\underline{0.211}  &\textbf{0.262}  
&\textbf{0.272} &\textbf{0.301} &\underline{0.373} &\textbf{0.458}  &\underline{0.644}  \\
\midrule
\end{tabular}
}
\end{table*}

\begin{table*}[]
\centering
\caption{The effect of of data augmentation for multivariate time series forecasting.}
\label{ablation_mpre}
\scalebox{0.55}
{
\begin{tabular}{ll|ccccc|ccccc|ccccc|ccccc}
\toprule
\multicolumn{2}{l|}{Datasets}        & \multicolumn{5}{c|}{ETTh1} & \multicolumn{5}{c|}{ETTh2} & \multicolumn{5}{c|}{ETTm1} & \multicolumn{5}{c}{Electricity} \\
\midrule
Methods & Metrics &24 &48 &168 &336 &720  &24 &48 &168 &336 &720   &24 &48 &96 &288 &672  &24     &48     &168     &336  &720  \\
\midrule
\multirow{2}{*}{\textbf{DE-TSMCL-pre}}     &$MSE(\downarrow)$ 
&\underline{0.581} &\underline{0.624} &\textbf{0.726} &\underline{0.907}  &1.049
     &\underline{0.396}  &\textbf{0.527}  &1.835 &2.155  &2.389 
  &{0.430} &{0.588}  &\underline{0.613} &0.703 &{0.780}  
      &\underline{0.257} &0.299 &\underline{0.308} &\underline{0.345}  &\underline{0.347}\\
& $MAE(\downarrow)$ 
&\underline{0.524} &\underline{0.551} &\underline{0.632} &\underline{0.713} &{0.794}
&\underline{0.459} &\underline{0.568} &1.061 &\underline{1.197} &1.291    
&{0.423} &{0.519} &0.551 &{0.601} &\underline{0.647}
&0.275 &0.306 &\textbf{0.372} &\underline{0.433} &\underline{0.441}\\
\midrule
\multirow{2}{*}{\textbf{DE-TSMCL-after}}     &$MSE(\downarrow)$ 
&{0.611} &0.644 &{0.773} &{0.914}  &\underline{1.043}   
&0.411 &{0.601} &\underline{1.821} &\underline{2.128} &\textbf{2.326}
&0.444 &{0.607} &{0.614} &{0.698} &0.798
&{0.269} &\underline{0.297} &0.330&{0.347} &0.370 \\
                          & $MAE(\downarrow)$ 
 &{0.537} &{0.564} &{0.649} &{0.718} &\underline{0.792}
 &{0.480} &{0.598} &\textbf{1.041} &\textbf{1.144} &\textbf{1.210}        
 &{0.428} &{0.521} &{0.543} &{0.601} &0.660     
 &\underline{0.274} &\underline{0.304} &0.394 &{0.450} &0.445  \\
\midrule

\multirow{2}{*}{\textbf{DE-TSMCL-reverse}}     &$MSE(\downarrow)$ 
&{0.623} &{0.631} &{0.769} &{0.909} &{1.052}    
&{0.419} &{0.595} &{1.903} &{2.138} &{2.412}     
&\underline{0.398} &\underline{0.565} &{0.614} &\underline{0.673}  &\underline{0.752}    
&{0.268} &{0.305} &{0.314} &{0.356} &{0.381}  \\
& $MAE(\downarrow)$ 
&{0.546} &{0.570} &{0.647} &{0.725} &{0.802}
 &{0.461} &{0.583} &{1.078} &{1.230}  &\underline{1.245}
 &\underline{0.404} &\underline{0.492} &\underline{0.513} &\underline{0.597}  &{0.659}  
 &{0.288} &{0.306} &{0.394} &{0.461} &{0.473}  \\
\midrule

\multirow{2}{*}{\textbf{DE-TSMCL}}     &$MSE(\downarrow)$ 
&\textbf{0.569} &\textbf{0.620} &\underline{0.744} &\textbf{0.899} &\textbf{1.002}    
&\textbf{0.376} &\underline{0.564} &\textbf{1.818} &\textbf{2.120} &\underline{2.376}     
&\textbf{0.391} &\textbf{0.549} &\textbf{0.601} &\textbf{0.660}  & \textbf{0.741}    
&\textbf{0.248} &\textbf{0.294} &\textbf{0.303} &\textbf{0.337} &\textbf{0.333}  \\
& $MAE(\downarrow)$ 
&\textbf{0.524} &\textbf{0.548} &\textbf{0.623} &\textbf{0.706} &\textbf{0.776}
 &\textbf{0.454} &\textbf{0.565} &\underline{1.058} &{1.213}  &{1.270}
 &\textbf{0.401} &\textbf{0.468} &\textbf{0.503} &\textbf{0.571}  &\textbf{0.611}  
 &\textbf{0.272} &\textbf{0.301} &\underline{0.373} &\textbf{0.408} &\textbf{0.424}  \\
\midrule
\end{tabular}
}\end{table*}

\subsection{Data Augmentation Analysis(RQ4)}
In this section, we provide a more comprehensive understanding of the effect of data augmentation on the model. In the design of our DE-TSMCL, data augmentation is performed after the projection layer for the teacher network, while before the projection layer for the student network. This hybrid approach differs from traditional distillation methods, which commonly adopt the same architecture for both the teacher and student networks.
Tables \ref{ablation_spre} and \ref{ablation_mpre} report the comparison results for univariate and multivariate time serial forecasting on four datasets. \textbf{Bold} represents the best performance, \underline{underline} represents the second best performance.
In particular, DE-TSMCL-pre utilizes augmentation before the projection layer for both the teacher and student networks, DE-TSMCL-after applies augmentation after the projection layer for both networks, and DE-TSMCL-reverse applies data augmentation before the projection layer for the teacher network, while after the projection layer for the student network. 

We observe that DE-TSMCL outperforms the other design in most cases across the evaluated datasets. The main reason is that applying data augmentation after the projection layer for the teacher network helps to introduce diversity and variability into the augmented samples. This can be beneficial for the teacher network's training, as it encourages robustness and better generalization by exposing it to a more diverse range of augmented samples.
On the other hand, implementing data augmentation before the projection layer for the student network allows the model to learn from the augmented samples with increased variability and complexity. This can help the student network to better capture and understand the augmented data, potentially leading to improved performance in terms of accuracy and generalization. Therefore, this evaluation provides insights into the optimal utilization of augmentation techniques in the distillation process.

\subsection{Statistical Significance Assessing}
In this section, we have incorporated the use of the Kolmogorov-Smirnov (K-S) test to assess the distributional similarity between the input and output sequences of various models. This statistical method helps determine whether the prediction results align with the distribution of the prediction sequence. The experiment is specifically conducted on the ETTm1 dataset with a fixed step size of 96, and the results of p-value are presented in Table\ref{ks}. The p-value in the context of K-S test represents the probability of observing a test statistic as extreme as the one calculated from the data. In the time series forecasting, a p-value less than a predetermined significance level (e.g 0.01) suggests that the observed difference between the input and output sequences is statistically significant. In our study, the p-values obtained from the K-S test are compared to the significance level of 0.01. When the p-values for the Transformer, Informer, and TS2Vec models are below 0.01, it suggests that the input and output sequences likely originate from different distributions. On the other hand, our proposed model yields a higher p-value, it indicates that the input and output sequences are likely to be from the same distribution. This finding further supports the superiority of our model in accurately capturing the distributional characteristics of the prediction sequence.

\begin{table*}[]
\centering
\caption{The K-S test results of different methods for univariate forecasting on ETTm1 dataset.}
\scalebox{1.0}{ 
\begin{tabular}{lccccccc}
\toprule
dataset      & \multicolumn{1}{c}{P} & \multicolumn{1}{c}{Transformer} & \multicolumn{1}{c}{Informer} & \multicolumn{1}{c}{TS2Vec}  & \multicolumn{1}{c}{DE-TSMCL}   \\ 
\midrule
           & 24   & 0.011  & 0.01 &0.014  &\textbf{0.035} \\  
           & 48   & 0.0097  & 0.0078 &0.0074 &\textbf{0.029}  \\  
ETTm1   & 96   &0.0061  & 0.0039 &0.0051 &\textbf{0.025} \\  
           & 288  &0.0022  & 0.0019 &0.0037 &\textbf{0.016} \\ %
           & 672  & 0.0023  & 0.0016 &0.0028 &\textbf{0.0094}  \\  

\bottomrule
\end{tabular}
}\label{ks}
\end{table*}

\section{Conclusion}\label{6}
\par 
In this paper, we propose a novel framework called Distillation Enhanced Time Series Network with Momentum Contrastive Learning (DE-TSMCL). DE-TSMCL incorporates knowledge distillation between models that utilize overlapping sub-series to represent time series data in forecasting tasks. Additionally, we leverage the advantages of two types of tasks: adaptive supervised task and self-supervised task, within the DE-TSMCL framework. This combination allows us to enhance the model's representation and improve its forecasting performance.
We conduct extensive experiments on five real-world datasets to evaluate the performance of DE-TSMCL. The results demonstrate that DE-TSMCL outperforms other state-of-the-art models in various scenarios. Notably, on the ETTm1 dataset, DE-TSMCL achieves a significant 24.2\% improvement in Mean Squared Error (MSE) and a 14.7\% improvement in Mean Absolute Error (MAE). Similarly, on the ETTh1 dataset, DE-TSMCL achieves substantial gains, with a 14.2\% improvement in MSE and a 10.6\% improvement in MAE.

We acknowledge that our proposed method has certain limitations. These limitations provide opportunities for future improvement and research. Some of the limitations of our proposed method include: (1) Scalability: Our method's performance may be affected when dealing with extremely large-scale time series datasets. The computational and memory requirements of the model may become a limiting factor. To address this, we plan to explore techniques such as model parallelism, distributed computing, or more efficient architectures to improve scalability.
(2) Generalization to diverse domains: Although our method demonstrates promising results in the specific domain we evaluated, its generalizability to diverse domains and datasets needs further investigation. We aim to conduct experiments on a wider range of datasets from different domains to assess the robustness and adaptability of our method.

In the future, our focus will be on exploring additional convolutional techniques, such as sparse convolution in 3D point clouds, as well as novel CNN architectures to enhance the capabilities of our model. We also aim to investigate more novel loss function approaches, such as Ranking-Based Cross-Entropy Loss and Triplet Loss to enhance the training process and improve the performance of the model. Furthermore, we plan to extend the applicability of our framework to other time-series analysis tasks.










\bibliographystyle{cas-model2-names}
\bibliography{ref}


\end{document}